\documentclass{article}

\usepackage{arxiv}

\usepackage[utf8]{inputenc} 
\usepackage[T1]{fontenc}    
\usepackage{hyperref}       
\usepackage{url}            
\usepackage{booktabs}       
\usepackage{amsfonts}       
\usepackage{nicefrac}       
\usepackage{microtype}      
\usepackage{graphicx}
\usepackage{natbib}
\usepackage{doi}

\usepackage{amsmath}
\usepackage{amsthm}
\usepackage[switch]{lineno}
\usepackage[table,usenames,dvipsnames]{xcolor}
\usepackage[small]{caption}
\usepackage{subcaption}
\usepackage{amssymb}
\usepackage{mathtools}
\usepackage{bm}
\usepackage{xspace}
\usepackage{url}
\usepackage{comment}

\usepackage{microtype}      
\usepackage{algorithm}
\usepackage[noend]{algorithmic}
\usepackage[inline]{enumitem}
\usepackage{comment}
\usepackage{threeparttable}
\usepackage{tikz}
\usepackage{multirow}
\usepackage{float}
\usepackage{placeins}
\usepackage{balance}

\title{NysAct: A Scalable Preconditioned Gradient Descent using Nystr\"om Approximation}

\author{Hyunseok Seung \\
	Department of Statistics\\
	University of Georgia\\
	Athens, GA \\
	\texttt{hseung@uga.edu} \\
	\And
	Jaewoo Lee \\
	School of Computing\\
	University of Georgia\\
	Athens, GA \\
	\texttt{jaewoo.lee@uga.edu} \\
	\And
	Hyunsuk Ko \\
	School of Electrical Engineering\\
	Hanyang University\\
	Ansan, South Korea \\
	\texttt{hyunsuk@hanyang.ac.kr} \\
}

\date{}

\renewcommand{\headeright}{}
\renewcommand{\undertitle}{}
\renewcommand{\shorttitle}{NysAct: A Scalable Preconditioned Gradient Descent using Nystr\"om Approximation}

\hypersetup{
  pdftitle={NysAct: A Scalable Preconditioned Gradient Descent using Nystrom Approximation},
  pdfsubject={cs.LG},
  pdfauthor={Hyunseok Seung},
  pdfkeywords={Deep learning optimization, Gradient preconditioning, Nystrom approximation}
}

\let\en=\ensuremath

\DeclareMathOperator{\E}{\mathbb{E}}          

\DeclareMathOperator{\Trace}{Tr}
\DeclareMathOperator{\diag}{diag}      

\DeclareMathOperator{\vect}{vec}                

\renewcommand{\vec}[1]{\en{\bm{\mathrm{#1}}}}

\newcommand{\mat}[1]{\en{{\bm{\mathrm{#1}}}}}
\newcommand{\grad}[0]{\en{\nabla}}

\newcommand{\R}[0]{\mathbb{R}}

\newcommand{\pdv}[2]{\frac{\partial #1}{\partial #2}}

\renewcommand{\th}[0]{\textsuperscript{th}\xspace}

\newcommand{\nysact}[0]{\textsc{NysAct}\xspace}

\theoremstyle{plain}
\newtheorem{theorem}{Theorem}[section]

\newtheorem{remark}[theorem]{Remark}
\theoremstyle{definition}

\newtheorem{assumption}[theorem]{Assumption}

\usepackage{pifont}
\newcommand{\xmark}{\ding{55}}%

\newcommand*\circled[1]{\tikz[baseline=(char.base)]{
            \node[shape=circle,draw,inner sep=0pt] (char) {#1};}}

\begin{document}
\maketitle

\begin{abstract}
	Adaptive gradient methods are computationally efficient and converge quickly, but they often suffer from poor generalization. In contrast, second-order methods enhance convergence and generalization but typically incur high computational and memory costs. In this work, we introduce \nysact, a scalable first-order gradient preconditioning method that strikes a balance between state-of-the-art first-order and second-order optimization methods. \nysact leverages an eigenvalue-shifted Nystr\"om method to approximate the activation covariance matrix, which is used as a preconditioning matrix, significantly reducing time and memory complexities with minimal impact on test accuracy. Our experiments show that \nysact not only achieves improved test accuracy compared to both first-order and second-order methods but also demands considerably less computational resources than existing second-order methods. Code is available at \href{https://github.com/hseung88/nysact}{https://github.com/hseung88/nysact}.


\end{abstract}


\vspace{1em}
This is the extended version of the paper published in the \textit{2024 IEEE International Conference on Big Data (BigData)}, \textcopyright\ IEEE. The published version is available at: \href{https://doi.ieeecomputersociety.org/10.1109/BigData62323.2024.10825352}{10.1109/BigData62323.2024.10825352}

\section{Introduction}
The success of deep learning models heavily depends on optimization
strategies, with gradient-based methods being crucial for effective
training. Gradient preconditioning has gained traction for its ability
to accelerate convergence by adjusting gradients during
training.
First-order methods such as stochastic gradient descent with
momentum (SGD)~\cite{robbins1951Stochastic}
and Adam(W)~\cite{kingma2015adam, Loshchilov2019DecoupledWD} are
popular for their computational efficiency, with Adam(W) using
adaptive learning rates based on the second moments of
gradients. However, despite  
%
their low per-iteration cost, their convergence is often slow.

Second-order methods can improve the convergence 
by 
preconditioning gradients
to  make them effective in navigating
ill-conditioned loss landscapes~\cite{Gupta2018ShampooPS,
  Goldfarb2020PracticalQM, Liu2024Sophia}. However, their
computational overhead is often prohibitive, especially in large-scale
deep learning tasks. For example, directly leveraging the Hessian
matrix~\cite{Tran2022Better, frangella2024sketchysgd} as a
preconditioner requires double backpropagation, significantly
increasing time and memory demands.
%
%
To improve efficiency, methods like KFAC~\cite{martens2015optimizing} approximates the (empirical)
Fisher information matrix (FIM), instead of the Hessian, and decompose
it into the Kronecker 
product of smaller matrices but still result in longer training times
compared to SGD. For example, in our experiments with ResNet-110 on
CIFAR-100 using a single GPU, AdaHessian~\cite{Yao2020ADAHESSIANAA} 
took on average 46.33 seconds per
epoch, Shampoo~\cite{Gupta2018ShampooPS} 200.63 seconds, and KFAC 21.75 seconds, while SGD took only 8.88 seconds.  


In this work, we propose a novel stochastic preconditioned optimizer
called \nysact, whose performance is as good as that of second-order
methods while requiring significantly less computation and memory.
KFAC approximates the FIM $\mat{F}\in \R^{mn \times mn}$ of a layer with
the Kronecker product of two matrices: $\mat{F} = \mat{A}\otimes
\mat{P}$, where $\mat{A}\in
\R^{n\times n}$ and $\mat{P}\in \R^{m\times m}$ are covariance matrix of activations and pre-activation 
gradients, respectively. A recent
work~\cite{Benzing2022GradientDO} 
empirically found that the term $\mat{P}$ makes no contribution to the high
performance of KFAC and proposed an optimizer, called FOOF, that
replaces $\mat{P}$ with an identity matrix $\mat{I}$.
Based on this observation,
\nysact chooses to use activation covariance matrix as a
preconditioner. While only computing and maintaining $\mat{A}$ saves
both computation and memory space, FOOF still suffers 
from low scalability due to the need for complex matrix operations on $\mat{A}$, 
rendering it impractical
for large neural networks.

To improve scalability, \nysact further
approximates the activation covariance matrix $\mat{A}$ using the eigenvalue-shifted Nystr\"om
method~\cite{Tropp2017FixedRankAO,Ray2021SublinearTA}.
%
Given a matrix $\mat{A} \in \R^{n\times n}$ and a fixed rank $r \ll n$, 
the Nystr\"om method 
requires only $\mathcal{O}(r^3 + n r^2)$ time and $\mathcal{O}(nr)$ memory to compute the
approximation of $\mat{A}^{-1}$. In contrast, low-rank approximations
that use singular value decomposition (SVD) have time and memory
complexities of $\mathcal{O}(n^3)$ and $\mathcal{O}(n^2)$,
respectively.
%
The eigenvalue-shifted Nystr\"om
method due to Tropp et al.~\cite{Tropp2017FixedRankAO} is specifically
adapted for positive semi-definite matrices and offers a sharp
approximation error bound, making it an effective choice for scalable gradient preconditioning in
deep learning.
We also make an important observation. There are two commonly used
sampling methods for Nyst\"om approximation, uniform column sampling
without replacement and Gaussian sketching. 
In our experiments comparing the efficacy of these two sampling methods 
for curvature matrix approximation,
we observed that both methods perform similarly when the input dataset is small to medium-scale dataset but, when the input is large-scale, the Gaussian sketching method 
performs worse than the subcolumn sampling method.
%

\begin{figure*}[tb]
    \centering
    \begin{subfigure}[t]{0.49\linewidth}
        \centering
        \includegraphics[width=\linewidth]{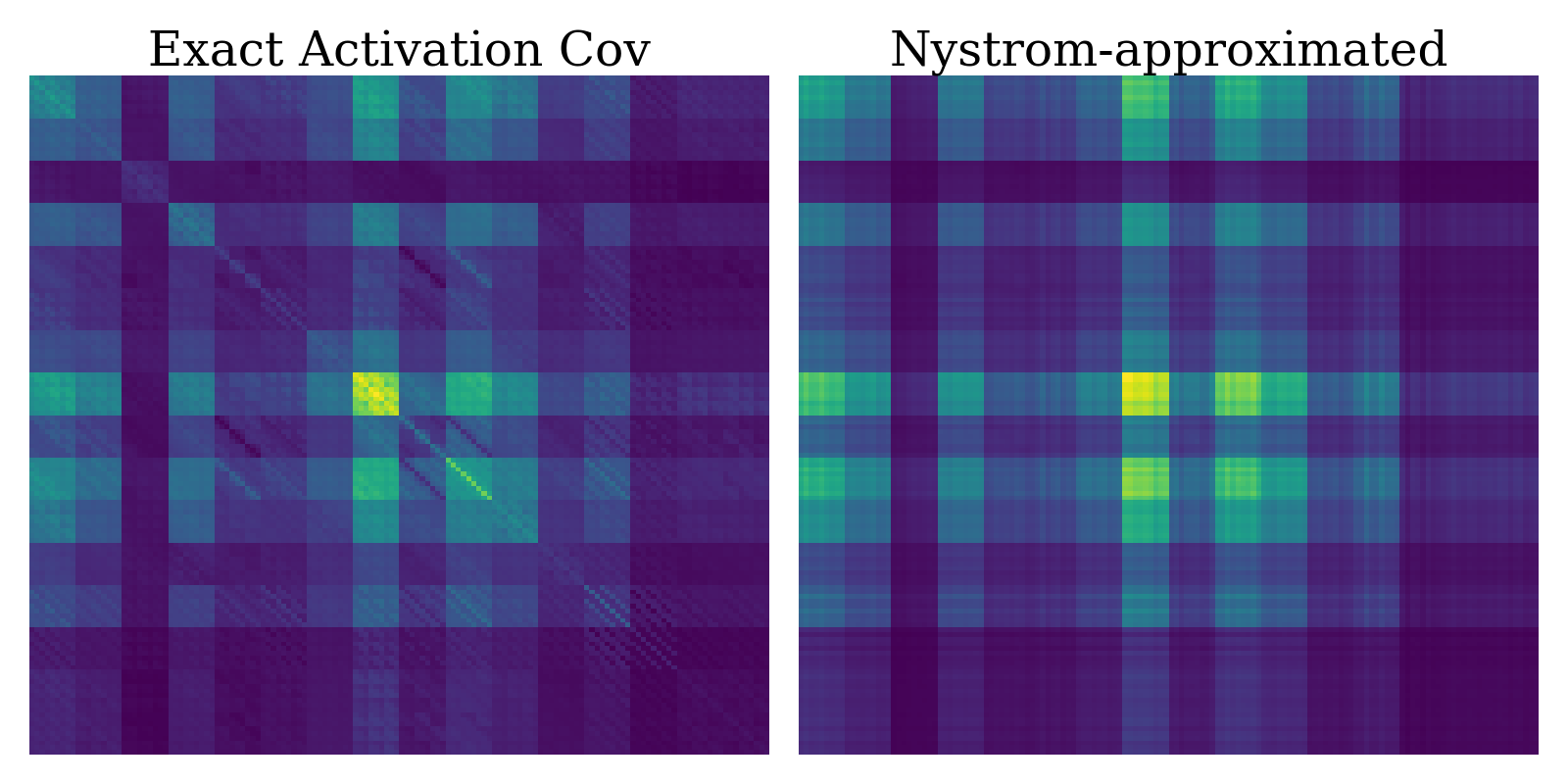}
        \caption{Layer1.2.Conv1}
    \end{subfigure}
    \hfill
    \begin{subfigure}[t]{0.49\linewidth}
        \centering
        \includegraphics[width=\linewidth]{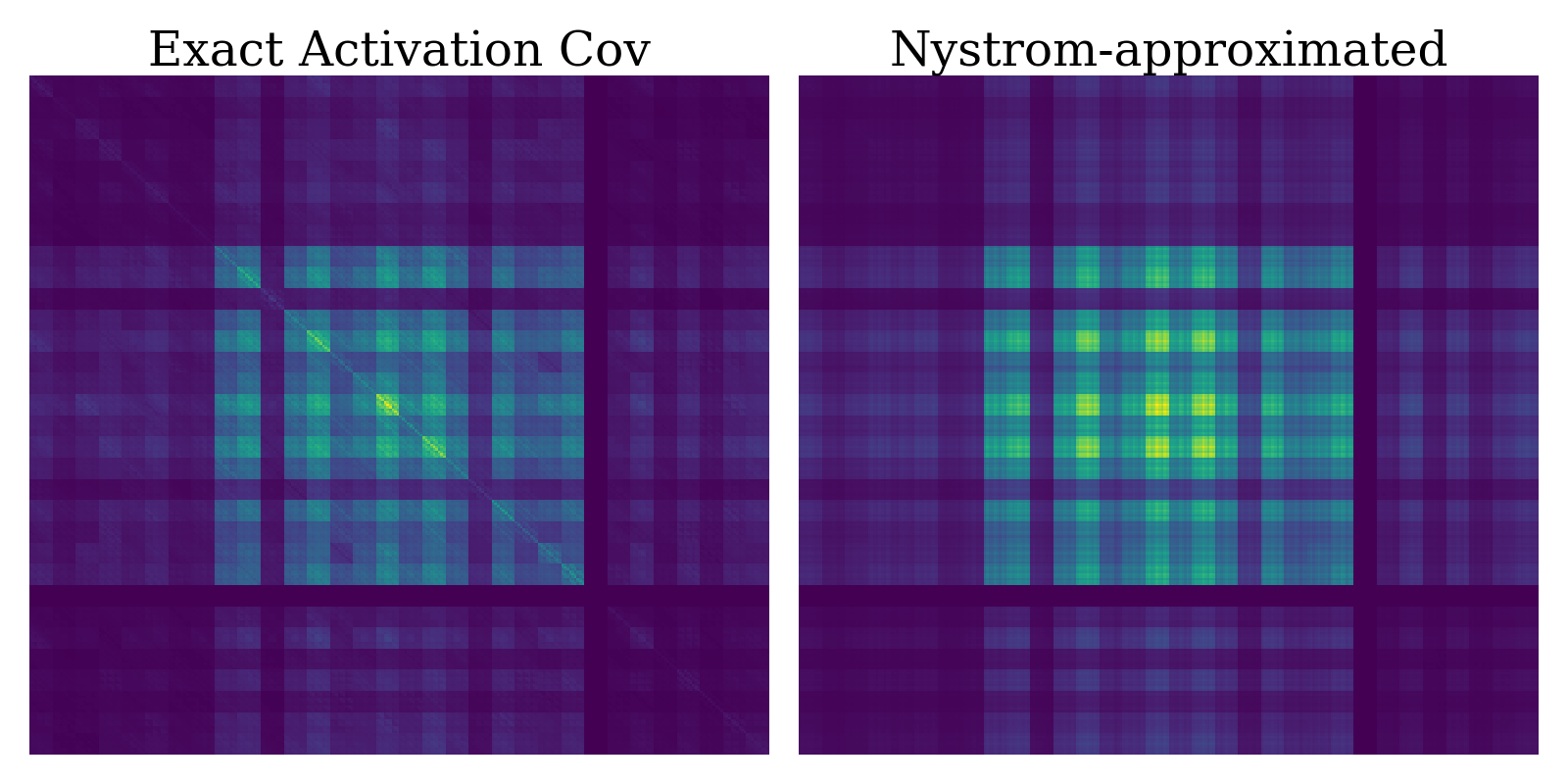}
        \caption{Layer2.1.Conv1}
    \end{subfigure}
    \caption{Comparison of exact activation covariance and Nyström-approximated activation covariance in ResNet-32 architecture trained on CIFAR-100 dataset}
    \label{fig:compare_actv}
\end{figure*}

To demonstrate the effectiveness of the Nystr\"om method in
approximating the activation covariance matrix, we trained
ResNet-32 model~\cite{he2016deep} on CIFAR-100~\cite{krizhevsky2009learning}
dataset for 100 epochs using SGD with a mini-batch size of
128. Figure~\ref{fig:compare_actv} shows the heatmaps of actual and
Nystr\"om approximated covariance matrices. As shown, the Nystr\"om
method has an ability to recover the whole covariance matrix from the randomly sampled
subset of $r$ columns.

\subsection{Contributions}
The key contributions of this paper are summarized as follows.

\textbf{Scalable Gradient Preconditioning. } We introduce \nysact, a
scalable gradient preconditioning method that significantly reduces
computational costs while maintaining a minimal compromise in
performance. By integrating the Nystr\"om approximation with
activation covariance, our method strikes a balance between efficiency
and accuracy, making it well-suited for large-scale deep learning
tasks. 

\textbf{Convergence Analysis.} We present a detailed convergence analysis of \nysact, showing that it achieves a convergence rate of $\mathcal{O}(1/T)$, thereby establishing theoretical guarantees for non-convex optimization problems. The analysis highlights how the Nystr\"om approximation impacts the convergence rate, offering valuable insights into the trade-offs between computational efficiency and optimization performance. This theoretical contribution provides a solid foundation for understanding the behavior of \nysact in deep learning applications.

\textbf{Extensive Experimental Validation. } We provide extensive
experimental results that demonstrate the effectiveness of \nysact in
image classification tasks. Our experiments, conducted across various
network architectures on CIFAR and ImageNet datasets, show that
\nysact not only achieves higher test accuracy than both state-of-the-art
first-order and second-order methods but also
requires less time and memory resources compared to second-order
methods.


\label{sec:intro}

\section{Related Work}
The work closest to ours is SketchySGD~\cite{frangella2024sketchysgd} which also uses the Nystr\"om method~\cite{Williams2000UsingTN} to obtain a preconditioner. Specifically, it approximates the (minibatch) Hessian with a low-rank 
matrix using Hessian vector products.
However, SketchySGD requires double backpropagation to compute these approximations, leading to substantial memory and time overhead. In our experiments of training ResNet-110 on CIFAR-100 dataset, SketchySGD required an average 58.14 seconds per epoch and used 21.3 GB of memory with a mini-batch size of 128 on a single GPU. In contrast, \nysact took only 11.32 seconds and 1.1 GB of memory. Although SketchySGD is most closely related to \nysact, we excluded it from our baselines due to 
its unacceptably high resource requirement.

Other notable methods that use a low-rank scalable approximation for preconditioning include GGT~\cite{Agarwal2020EfficientFA}, K-BFGS~\cite{Goldfarb2020PracticalQM}, MFAC~\cite{frantar2021mfac}, SKFAC~\cite{Tang2021SKFACTN}, and Eva~\cite{Zhang2023EvaPS}. 
GGT exploits the low-rank structure of the sum of the outer product of the gradient and efficiently computes its inverse square root using SVD.
While it has lower computational overhead compared to full-matrix preconditioning, 
it remains resource-intensive 
because of the need for performing SVD on the matrix formed by past gradients.
K-BFGS efficiently computes the inverse of the approximated FIM used in KFAC by employing the BFGS method~\cite{Broyden1970TheCO, Fletcher1970ANA, Goldfarb1970AFO, Shanno1970ConditioningOQ}. However, it requires additional forward and backward passes to compute the $(\vec{s}, \vec{y})$ pair for BFGS updates, increasing the overall computational cost. MFAC introduced rank-1 approximations for estimating inverse-Hessian vector products, employing iterative conjugate gradient solvers. However, this method necessitates multiple forward and backward passes, which significantly raises both computational and memory requirements. SKFAC proposed a low-rank formulation for the inverse of FIM using the Sherman-Morrison-Woodbury formula. It stores both the activation covariance and pre-activation gradient covariance matrices as used in KFAC, requiring the inversion of both matrices, whereas \nysact stores only sketched activation covariance matrices. Among KFAC's low-rank approximation variants, Eva is the most efficient method that preserves KFAC's original performance. Eva computes and stores batch-averaged activation and pre-activation gradient vectors, and updates the inverse of the approximated FIM using the Sherman-Morrison formula. However, as demonstrated in \cite{Benzing2022GradientDO}, KFAC's effectiveness as a second-order method is primarily driven by the activation term, rather than the pre-activation gradient term.
\label{sec:related}

\section{Preliminaries}
\label{sec:prelim}
\subsection{Notations}
We use lower-case bold letters for (column) vectors and upper-case
bold letters for matrices. $\mat{I}_m$ is an $m\times m$ identity matrix.
For a matrix
$\mat{M} \in \R^{p \times q}$, its vectorization, denoted by
$\vect(\mat{M})$, is the column vector of size $pq\times 1$ obtained
by concatenating columns of $\mat{M}$, i.e., 
$\vect(\mat{M}) = \begin{bmatrix} 
\mat{M}_{*,1}^{\intercal} & \mat{M}_{*,2}^{\intercal} & \cdots & \mat{M}_{*,q}^{\intercal}
\end{bmatrix}^{\intercal}\,$, where $\mat{M}_{*,j}$ denotes the $j$\th
column of matrix $\mat{M}$.
For any matrix $\mat{M}$, we denote the set of its eigenvalues by
$\lambda(\mat{M})$ and the set of its singular values by
$\sigma(\mat{M})$, both assumed to be sorted in descending order. The Frobenius norm is denoted by $\|\mat{M}\|_F$, and the Kronecker product is represented by $\otimes$. The set $\{1, 2, \dots, N\}$ is denoted by $[N]$. 

\subsection{Setup for Architecture and Training}
Consider a network $f(\vec{x};\vec{\theta})$ composed of $L$ layers,
trained on a dataset $\mathcal{D} = \{(\vec{x}_i, y_i)\}_{i=1}^n$. For
each layer $l \in [L]$, let $\mat{W}^{(l)}\in \R^{d_l\times d_{l-1}}$
represents the weight matrix, and $\vec{b}^{(l)} \in \R^{d_l}$
represents the bias vector. The forward 
step of $f$ is defined as follows: 
\begin{align*}
    &\vec{z}^{(l)} = \mat{W}^{(l)} \vec{a}^{(l-1)} + \vec{b}^{(l)}\in \R^{d_{l}}\,, \\
    &\vec{a}^{(l)} = \phi(\vec{z}^{(l)})  \in \R^{d_{l}}\,,
    \qquad \vec{a}^{(0)} = \vec{x}\,,\\
    &\vec{\theta}^{(l)} = [\vect(\mat{W}^{(l)})^{\intercal}, (\vec{b}^{(l)})^{\intercal}]^{\intercal} \in
    \R^{d_{l}(d_{l-1}+ 1)}\,, \\
    &\vec{\theta} \in [(\vec{\theta}^{(1)})^{\intercal}, \ldots,
    (\vec{\theta}^{(L)})^{\intercal}]^{\intercal}\in \R^{p}\,,
\end{align*}
where $\vec{z}$ denotes the pre-activations, $\vec{a}$ represents the activations, and $\phi$ is the activation function.
For convolutional layers, similar to KFAC, we employed patch
extraction to unfold the input into patches, transforming the
additional axes into a format compatible with matrix operations. 

We consider training a deep neural network $f:\R^d \to \R$ that takes
an input $\vec{x}$ and produces an output $f(\vec{\theta};\vec{x})$.
Given training examples $\mathcal{D}$, we aim to learn the network 
parameters $\vec{\theta}$ by minimizing the empirical loss
$\mathcal{L}$ over the training set:  
\begin{equation} \label{eq:opt_prob}
  \min_{\vec{\theta} \in \R^{d}} \mathcal{L}(\vec{\theta}; \mathcal{D}) :=
  \frac{1}{n}\sum_{i=1}^{n}{\ell\left(f(\vec{x}_i;\vec{\theta}), y_i
    \right)}, 
\end{equation}
where $\ell$ is a loss function. Throughout the paper, $\mathcal{B}_k$
denotes the minibatch at iteration $k$ constructed by randomly sampling
examples in $\mathcal{D}$.

\subsection{FIM-based Gradient Preconditioning}
%
To solve the problem~\eqref{eq:opt_prob}, KFAC approximates the FIM with a Kronecker product of smaller matrices as $\mat{F}^{(l)} \approx \mat{A}^{(l-1)}\otimes \mat{P}^{(l)}$, where $\mat{A}^{(l)}=\E\left[\vec{a}^{(l)}(\vec{a}^{(l)})^\intercal\right]$ denotes the activation covariance from layer $l$, and $\mat{P}^{(l)}=\E\left[\pdv{\mathcal{L}}{\vec{z}^{(l)}} \pdv{\mathcal{L}}{\vec{z}^{(l)}}^\intercal\right]$ represents the pre-activation gradient covariance from layer $l$.

Assuming the independence between layer $i$ and $j$ for $i\neq j$,
KFAC computes the diagonal blocks of FIM only, which results in the
following update rule for layer $l$ at iteration $k$. 
\begin{align}
  \label{eq:kfac}
  \vec{\theta}_{k+1}^{(l)}
  &=\vec{\theta}_{k}^{(l)} -\eta_k (\mat{A}_k^{(l-1)} \otimes \mat{P}_k^{(l)})^{-1} \vec{g}_{k}^{(l)} \nonumber \\ 
    &=\vec{\theta}_{k}^{(l)}-\eta_k
    \vect{\left((\mat{P}_{k}^{(l)})^{-1}\mat{G}_{k}^{(l)}(\mat{A}_{k}^{(l-1)})^{-1}\right)}\,,
\end{align}
where $\mat{G}$ represents the gradient of the loss with respect to the parameters, and $\vec{g}$ denotes the gradient in vectorized form. A notable scalable KFAC variant recently proposed is Eva which has following update rule:
\begin{align}
  \label{eq:eva}
  \vec{\theta}_{k+1}^{(l)}
    &=\vec{\theta}_{k}^{(l)}-\eta_k
    \vect{\left((\tilde{\mat{P}}_{k}^{(l)})^{-1}\mat{G}_{k}^{(l)}(\tilde{\mat{A}}_{k}^{(l-1)})^{-1}\right)}\,, \nonumber
\end{align}
where the matrix $\tilde{\mat{A}}$ is defined as $\tilde{\mat{A}} = \E\left[\vec{a}^{(l-1)}\right] \E\left[\vec{a}^{(l-1)}\right]^\intercal$ and the matrix $\tilde{\mat{P}}$ is given by $\tilde{\mat{P}} = \E\left[\pdv{\mathcal{L}}{\vec{z}^{(l)}}\right] \E\left[ \pdv{\mathcal{L}}{\vec{z}^{(l)}}\right]^\intercal$. The update rule for FOOF is given by substituting $\mat{P}$ in \eqref{eq:kfac} into the identity matrix $\mat{I}$:
\begin{equation}
  \label{eq:foof}
  \vec{\theta}_{k+1}^{(l)}=\vec{\theta}_{k}^{(l)}-\eta_k
  \vect{\left(\mat{G}_{k}^{(l)}(\mat{A}_{k}^{(l-1)})^{-1}\right)}\,. \nonumber
\end{equation}

\subsection{Nystr\"om Method}
The Nystr\"om method~\cite{Williams2000UsingTN} is a well-established
technique for constructing low-rank approximations of a matrix
$\mat{A}\in \R^{n\times n}$ by selecting a subset of its
columns. Specifically, 
let $\mat{S}\in \R^{n\times r}$ be a matrix that randomly samples
$r\ll n$ columns of $\mat{A}$, where each column of $\mat{S}$ is a
vector having one entry equal to 1 and all other entries are 0. Then
$\mat{A}\mat{S}\in \R^{n\times r}$ corresponds to the submatrix of
$\mat{A}$ formed by $r$ randomly sampled columns of $\mat{A}$. The
Nystr\"om approximation $\mat{A}_{\text{nys}}$ of $\mat{A}$ is given by
\[
  \mat{A}\approx \mat{A}_{\text{nys}} =  \mat{A}\mat{S}
  \mat(\mat{S}^\intercal\mat{A}\mat{S})^\dagger\mat{S}^\intercal\mat{A}\,,
\]
where $\mat{X}^\dagger$ denotes the Moore-Penrose
pseudoinverse. Notice that $\mat{A}_{\text{nys}}$ can be obtained by
storing $\mat{S}^\intercal\mat{A}\mat{S} \in \R^{r\times r}$ and
$\mat{A}\mat{S}\in \R^{n\times r}$, which only takes $\mathcal{O}(nr)$
memory space.

There are alternative ways of constructing the sampling matrix
$\mat{S}$. One way is to sample each entry of $\mat{S}$ from the
standard Guassian distribution $\mathcal{N}(0,1)$, which is called
Gaussian sketching. Instead of sampling columns, the Gaussian
sketching randomly projects the points in $\mat{A}$ onto a lower
dimensional space.

\section{Algorithm}
\label{sec:algo}
In this section, 
we describe the details of each step in the proposed \nysact algorithm.
The pseudocode 
is provided in Algorithm~\ref{alg:pseudocode}.

\begin{algorithm}[tb]
\caption{\nysact}
\label{alg:pseudocode}
\textbf{Require}: Learning rate $\eta_k$, Momentum $\beta_{1}$, EMA $\beta_{2}$, Damping $\rho$, Covariance update frequency $\tau_{cov}$, Inverse update frequency $\tau_{inv}$, Rank $r$ \\
\textbf{Initialize}: Parameter $\vec{\theta}_{0}^{(l)}$, Momentum vectors $\vec{m}_{0}^{(l)} = \vec{0}$, Sketching matrices $\Tilde{\mat{C}}_0^{(l)} = \mat{O}$, Preconditioning matrices $(\mat{C}_{0, \text{nys}}^{(l)})^{-1} = \mat{I}$
\begin{algorithmic}[1]
  \FOR{$k = 1, 2, 3, \dots$}
    \STATE Construct minibatch $\mathcal{B}_k$ by randomly sampling
    from $\mathcal{D}$
    \IF{Sketch == "Gaussian"}
    \STATE $\mat{S}^{(l)}=(S_{ij})_{1\leq i\leq d_{l-1},\, 1\leq j\leq
      r},\, S_{ij}\sim \mathcal{N}(0, \frac{1}{d_{l-1}})$ \label{alg:g_sample}
    \ELSIF{Sketch == "Subcolumn"}
        \STATE Indices = RandPerm$(d_{l-1})[:r]$
        \STATE $\mat{S}^{(l)} = \mat{I}_{d_{l-1}}{[\text{Indices}]}
        \in \R^{d_{l-1} \times r}$
        \label{alg:sub_sample}
    \ENDIF
    \IF{$(k \mod \tau_{cov}) = 0$}
        \STATE $\widetilde{\mat{C}}_k^{(l)} = \beta_{2} \cdot \widetilde{\mat{C}}_{k-1}^{(l)} + (1 - \beta_{2}) \cdot \mat{A}_{k}^{(l)}\mat{S}^{(l)} \in \R^{d_{l-1} \times r}$ \label{alg:cov_update}
    \ENDIF
    \IF{$(k \mod \tau_{inv}) = 0$}
        \STATE $\widehat{\mat{C}}_k^{(l)} = \widetilde{\mat{C}}_k^{(l)} / (1 -
        \beta_{2}^{\lfloor k / \tau_{cov}\rfloor})$ \label{alg:bias_correction}
        \STATE $\widehat{\mat{C}}_{k, \text{damped}}^{(l)} =\widehat{\mat{C}}_k^{(l)} + \rho \cdot \mat{S}^{(l)}$ \label{alg:damped_sketch}
        \STATE $\mat{W}^{(l)} = (\mat{S}^{(l)})^\intercal\widehat{\mat{C}}_{k, \text{damped}}^{(l)} \in \R^{r \times r}$ \label{alg:principal_submatrix}
        \STATE Eigendecomposition: $\mat{W}^{(l)}= \mat{Q}\mat{\Lambda}\mat{Q}^\intercal$ \label{alg:eigendecomp}
        \STATE $\mat{W}_{\text{shifted}}^{(l)} = \mat{Q}\left[\mat{\Lambda} + \left(|\lambda_{\min}(\mat{W}^{(l)})| + \rho\right)\mat{I}\right]\mat{Q}^\intercal$ \label{alg:eigenvalue_shift}
        \STATE Cholesky decomposition: $\mat{W}_{\text{shifted}}^{(l)}
        = \mat{R}^{\intercal}\mat{R}$ \label{alg:cholesky_decomp}
        \STATE $\mat{X}^{(l)} = \widehat{\mat{C}}_{k, \text{damped}}^{(l)}\mat{R}^{-1} \in \R^{d_{l-1} \times r}$ \label{alg:solve_triangular}
        \STATE Singular value decomposition: $\mat{X}^{(l)} = \mat{U}\mat{\Sigma}\mat{V}^\intercal$  \label{alg:svd}
        \STATE $\widetilde{\mat{\Sigma}} = \diag\left(\max\left(\vec{\sigma}^{2} - (|\lambda_{\min}(\mat{W}^{(l)})| + \rho)\vec{1}_r, \vec{0}_{r} \right)\right)$ \\
        \quad $\in \R^{r\times r}$ \label{alg:back_shift}
        \STATE $(\mat{C}_{k,\text{nys}}^{(l)})^{-1} = \mat{U}_{[:r]}\widetilde{\mat{\Sigma}}^{-1}\mat{U}_{[:r]}^\intercal + \frac{1}{\rho}(\mat{I} - \mat{U}_{[:r]}\mat{U}_{[:r]}^\intercal)$ \\
        \qquad \qquad$\in \R^{d_{l-1}\times d_{l-1}}$ \label{alg:inv_update}
    \ENDIF
    \STATE $\mat{G}_k = \nabla \mathcal{L}(\vec{\theta}_k) \in \R^{d_{l}\times d_{l-1}}$
    \STATE $\vec{m}_{k}^{(l)} = \beta_{1} \vec{m}_{k-1}^{(l)} - \eta_{k}\vect{\left(\mat{G}_{k}^{(l)}(\mat{C}_{k,\text{nys}}^{(l)})^{-1} \right)}$
    \STATE $\vec{\theta}_{k}^{(l)} = \vec{\theta}_{k-1}^{(l)} +
    \vec{m}_{k}^{(l)}$ \label{alg:param_update}
\ENDFOR
\end{algorithmic}
\end{algorithm}

At iteration $k$, the algorithm estimates the covariance
$\mat{A}_k^{(l)}$ of activations $\vec{a}^{(l)}$ for each layer $l$
using the examples in the minibatch $\mathcal{B}_k$: 
\[
  \mat{A}_k^{(l)} =
  \E\left[\vec{a}^{(l)}\left(\vec{a}^{(l)}\right)^{\intercal}\right]
  \approx \frac{1}{|\mathcal{B}_k|}\sum_{i\in \mathcal{B}_k}
  \vec{a}_i^{(l)}(\vec{a}_i^{(l)})^{\intercal}\,,
\]
where $\vec{a}_i^{(l)}$ denotes the activations of
network $f(\vec{x}_i;\vec{\theta}_k)$ at layer $l$.
Let $\mat{G}_k^{(l)} = \grad_{\vec{\theta}^{(l)}}\mathcal{L}(\vec{\theta}_k; \mathcal{B}_k) 
\in \R^{d_{l}\times d_{l-1}}$ be the minibatch-estimated
  gradient of layer $l$ with input  and output dimensions
  $d_{\text{in}}=d_{l-1}$ and $d_{\text{out}}=d_l$, respectively.
FOOF updates the parameters of layer $l$ as follows:
\begin{align*}
  \vec{\theta}_{k+1}^{(l)}
  &= \vec{\theta}_k^{(l)} -\eta_k(\mat{C}_k^{(l)} \otimes
    \mat{I}_{d_l})^{-1}\vect(\mat{G}_l^{(l)}) \\
  &=\vec{\theta}_k^{(l)} -\eta_k\vect\left(\mat{G}_k^{(l)}(\mat{C}_k^{(l)})^{-1}\right)\,,
\end{align*}
where $\mat{C}_k^{(l)}$ is the exponential moving average (EMA) of
covariance matrix of activations of $l-1$, given by
\begin{equation} \label{eq:foof_update}
  \mat{C}_k^{(l)} = \beta_2\mat{C}_{k-1}^{(l)} +
  (1-\beta_2)\mat{A}_k^{(l-1)} \in \R^{d_{\text{in}}\times d_{\text{in}}}\,.
\end{equation}
Compared to KFAC update in~\eqref{eq:kfac}, the
update in~\eqref{eq:foof_update} only requires computing and storing
the EMA of $\mat{A}_k^{(l)}$, which reduces time and memory complexity from
$\mathcal{O}(d_{\text{out}}^{3}+d_{\text{in}}^{3})$ to
$\mathcal{O}(d_{\text{in}}^{3})$ and from
$\mathcal{O}(d_{\text{out}}^{2}+d_{\text{in}}^{2})$ to
$\mathcal{O}(d_{\text{in}}^{2})$, respectively.
However, for large-scale networks, $\mathcal{O}(d_{\text{in}}^{3})$
can still be prohibitively expensive. To mitigate the issue, we
propose to obtain a randomized low-rank approximation
$\mat{C}_{k,\text{nys}}^{(l)}$ of  $\mat{C}_k^{(l)}$
using the Nystr\"om method. 

\subsection{Eigenvalue-shifted Nystr\"om}
%
%
Let $\mat{S}_k^{(l)}\in \R^{d_{\text{in}}\times r}$ be a sampling
matrix such that 
$\mat{A}_k^{(l)}\mat{S}_k^{(l)}$ is a random subset of $r$ columns
from $\mat{A}_k^{(l)}$ (line~\ref{alg:sub_sample}) or a Gaussian
sketching of $\mat{A}_k^{(l)}$ (line~\ref{alg:g_sample}). We call our
method \textsc{NysAct-S} when $\mat{S}_k^{(l)}$ samples the columns
and \textsc{NysAct-G} when $\mat{S}_k^{(l)}$ performs Gaussian
sketching. Instead of storing $\mat{A}_k^{(l)}\in \R^{d_{\text{in}}\times
  d_{\text{in}}}$, \nysact maintains EMA of
$\mat{A}_k^{(l)}\mat{S}_k^{(l)} \in\R^{d_{\text{in}}\times r}$, which
modifies the EMA update in~\eqref{eq:foof_update} as follows:
\begin{equation} \label{eq:nysact_update}
  \widetilde{\mat{C}}_{k}^{(l)}
  = \beta_2\widetilde{\mat{C}}_{k-1}^{(l)} + (1-\beta_2)\mat{A}_k^{(l-1)}\mat{S}_k^{(l)}\,.
\end{equation}
This greatly reduces the memory complexity to
$\mathcal{O}(d_{\text{in}}\cdot r)$, where the rank $r$ is normally
much smaller than the layer's input size $d_{\text{in}}$. In our experiments on CIFAR datasets, we set $r=10$.  
In line~\ref{alg:damped_sketch}, to ensure the positive
semi-definiteness $\mat{C}_k^{(l)}$, \nysact computes the sketch of 
damped $\mat{A}_k^{(l)}$ after applying the bias correction
(line~\ref{alg:bias_correction}) to $\widetilde{\mat{C}}_k^{(l)}$:
\[
  \widehat{\mat{C}}_{k,\text{damped}}^{(l)} =
  \widehat{\mat{C}}_{k}^{(l)}+\rho\mat{S}_k^{(l)} \approx (\mat{A}_k
  ^{(l-1)}+\rho\mat{I}_{d_{l-1}})\mat{S}_k^{(l)}\,,
\]
where $\rho>0$ is a damping factor.

For clarity, hereafter we omit the layer index $(l)$.
Our goal is to obtain the inverse of (damped) activation covariance
matrix $\mat{C}_{k}$ from the Nystr\"om sketch $\widetilde{\mat{C}}_{k}$.
One can use the standard  
Nystr\"om method~\cite{Williams2000UsingTN} along with the matrix
inversion lemma.
\begin{align}
  \mat{C}_k^{-1}
  &\approx \left(\rho\mat{I}_{d_{l-1}} +
    \widetilde{\mat{C}}_k(\mat{S}_k^{\intercal}\widetilde{\mat{C}}_k)^{\dagger}
    \widetilde{\mat{C}}_k^{\intercal}\right)^{-1} \nonumber \\
  &=\frac{1}{\rho}\mat{I}_{d_{l-1}}-\frac{1}{\rho^{2}}\widetilde{\mat{C}}_k\left(
    \mat{S}_k^{\intercal}\widetilde{\mat{C}}_k +
    \frac{1}{\rho}\widetilde{\mat{C}}_k^{\intercal}\widetilde{\mat{C}}_k
    \right)^{-1}\widetilde{\mat{C}}_k^{\intercal} \label{eq:woodbury}
\end{align}
However, the above equation is not numerically stable~\cite{frangella2023Randomized, Goldfarb2020PracticalQM}.
We also empirically observed that computing the inverse using~\eqref{eq:woodbury}
results in training instability.
Instead, we propose to use the numerically stable eigenvalue-shifted Nystr\"om
method~\cite{Tropp2017FixedRankAO, Ray2021SublinearTA}. The low-rank
approximation procedure presented in 
Algorithm~\ref{alg:pseudocode} combines the ideas
of~\cite{Tropp2017FixedRankAO} and~\cite{Ray2021SublinearTA} and adapt it
to deep learning settings.
To obtain a rank-$r$ approximation $\widehat{\mat{C}}_{k,\text{nys}}$ of
$\mat{C}_{k} +\rho\mat{I}_{d_{\text{in}}}$, \nysact computes the Cholesky
decomposition $\mat{S}_k^{\intercal} \widehat{\mat{C}}_{k,\text{damped}} =
\mat{R}^{\intercal}\mat{R} \in \R^{r\times r}$
in line~\ref{alg:cholesky_decomp}, which allows to express 
\begin{align*}
  \widehat{\mat{C}}_{k,\text{nys}}
  &=
  \widehat{\mat{C}}_{k,\text{damped}}(\mat{S}_k^{\intercal}
  \widehat{\mat{C}}_{k,\text{damped}})^{-1}\widehat{\mat{C}}_{k,\text{damped}}\\
  &=
  (\widehat{\mat{C}}_{k,\text{damped}}\mat{R}^{-1})(\widehat{\mat{C}}_{k,\text{damped}}\mat{R}^{-1})^{\intercal}
    = \mat{X}\mat{X}^{\intercal}\,.
\end{align*}
Let the eigenvalue decomposition of $\mat{X}$ be
given by $\mat{X}=\mat{U}\mat{\Sigma}\mat{V}^{\intercal}$. Then 
we have
$\widehat{\mat{C}}_{\text{nys}}=\mat{U}\mat{\Sigma}^{2}\mat{U}^{\intercal}$
and its inverse $\widehat{\mat{C}}_{k,\text{nys}}^{-1}$ can be easily
obtained by replacing the diagonal 
elements of $\mat{\Sigma}^{2}$ by their reciprocals
(line~\ref{alg:inv_update}). For numerical stability of linear
algebraic operations that requires positve-definite matrices,
line~\ref{alg:eigenvalue_shift} shifts the eigenvalues of $\mat{W}$
and they are shifted back in line~\ref{alg:back_shift} to remove the
effect of the shift..

\begin{table}[tb]
  \caption{Comparison of time and memory complexity for updating preconditioner(s)}
  \label{tab:complexity}
  \centering
  \begin{tabular}{lll}
    \toprule
    Method  & Time Complexity & Memory Complexity \\
    \midrule
    \textsc{KFAC} & $\mathcal{O}(d_{\text{out}}^3) + \mathcal{O}(d_{\text{in}}^3)$ & $\mathcal{O}(d_{\text{out}}^2) + \mathcal{O}(d_{\text{in}}^2)$ \\
    \textsc{Eva} & $\mathcal{O}(d_{\text{out}}^2) + \mathcal{O}(d_{\text{in}}^2)$ & $\mathcal{O}(d_{\text{out}}) + \mathcal{O}(d_{\text{in}})$ \\
    \midrule
     \textsc{FOOF} & $\mathcal{O}(d_{\text{in}}^3)$ & $\mathcal{O}(d_{\text{in}}^2)$ \\
    \nysact & $\mathcal{O}(r^3) + \mathcal{O}(d_{\text{in}}\cdot r^2)$ & $\mathcal{O}(d_{\text{in}}\cdot r)$ \\
    \bottomrule
  \end{tabular}
\end{table}

\subsection{Complexities}
We compare the asymptotic time and memory costs of preconditioning a
layer with a weight matrix of size $(d_{\text{out}} \times
d_{\text{in}})$ in Table~\ref{tab:complexity}. The most
computationally intensive steps in \nysact occur in
Line~\ref{alg:eigendecomp}, \ref{alg:cholesky_decomp},
\ref{alg:solve_triangular}, and \ref{alg:svd}. However, since the rank
$r$ is typically much smaller than the dimensions of the weight
matrix, \nysact is expected to have 
significantly lower complexity than
KFAC and FOOF, 
making it more scalable and suitable for large-scale deep learning
tasks.

\subsection{Hyperparameters}
%
%
\nysact is robust to variations in the learning rate $\eta$, EMA coefficient $\beta_2$, damping factor $\rho$, and inverse update frequency $\tau_{inv}$, allowing for
consistent and reliable performance with less effort in hyperparameter
tuning. In our experiments, we observed that \nysact, FOOF, and other
KFAC variants perform well when using the same hyperparameters as SGD,
such as the learning rate, momentum coefficient $\beta_1$, and weight
decay. 

\section{Convergence Analysis}
\label{sec:analysis}
In this section, we analyze the convergence properties of \nysact. To
simplify the analysis, we focus on feed-forward networks composed of
linear layers, though these results can easily be extended to other types of
layers as well. We make the following common standard assumptions 
in stochastic optimization. 

\begin{assumption}[Smoothness] \label{as:smoothness}
The loss function $\mathcal{L}$ is continuously differentiable and $L$-smooth, i.e., for all $\mat{W}_1, \mat{W}_2 \in \R^{m \times n}$, 
\begin{equation*}
    \|\nabla\mathcal{L}(\mat{W}_1) - \nabla\mathcal{L}(\mat{W}_2)\|_F \leq L\|\mat{W}_1 - \mat{W}_2\|_F
\end{equation*}
\end{assumption}
\begin{assumption}[Gradient properties] \label{as:grad_prop}
  We make the following assumptions about the
  stochastic gradient $\mat{G}(\mat{W})$:
  \begin{enumerate}[label=(\roman*),leftmargin=*,noitemsep]
  \item $\mat{G}(\mat{W})$ is an unbiased estimate of
    the true gradient $\grad{\mathcal{L}}(\mat{W})$, i.e., 
    \begin{equation*}
      \E[\mat{G}(\mat{W})] = \nabla\mathcal{L}(\mat{W}).
    \end{equation*}
  \item The variance of $\mat{G}(W)$ 
    is bounded by $\sigma^2>0$:
    \begin{equation*}
      \E\left[\|\mat{G}(\mat{W}) - \nabla\mathcal{L}(\mat{W})\|_{F}^2\right] \leq \sigma^2.
    \end{equation*}
  \item The second moment of the true gradient is bounded by a constant $C$:
    \begin{equation*}
      \E\left[\|\nabla\mathcal{L}(\mat{W})\|_F^2\right] \leq C.
    \end{equation*}    
  \end{enumerate}
\end{assumption}
Further, we make the following assumptions to guarantee that the
activation covariance matrix remains 
well bounded and the
Nystr\"om approximation error is properly controlled. 
\begin{assumption}[Bounded Activation Covariance] \label{as:bounded_actv}
The Frobenius norm of activation covariance $\mat{A} \in \R^{d_{\text{in}} \times d_{\text{in}}}$ is bounded by constants $0 < \nu_L < \nu_U$, such that
\begin{equation*}
    \nu_L \leq \|\mat{A}\|_F \leq \nu_U. 
\end{equation*}
Also, the eigenvalues of $\mat{A}$ are bounded by constants $0 < \lambda_{L} < \lambda_{U}$, such that, for all $i\in [d_{\text{in}}]$ and $\mathcal{B} \subset \mathcal{D}$,
\begin{equation*}
    \lambda_{L} < \lambda_{i}(\mat{A}_{\mathcal{B}}) < \lambda_{U}.
\end{equation*}
\end{assumption}
\begin{assumption}[Nystr\"om Approximation] \label{as:nystrom_error}
The Nystr\"om approximated activation covariance $\mat{A}_{nys}$ of the exact activation covariance $\mat{A}$ satisfies the approximation bound
\begin{equation*}
    \|\mat{A} - \mat{A}_{nys}\|_F \leq \epsilon\|\mat{A}\|_F
\end{equation*}
for some small approximation error $\epsilon > 0$. 
\end{assumption}

\begin{figure}[tb]
    \centering
    \includegraphics[width=0.7\linewidth]{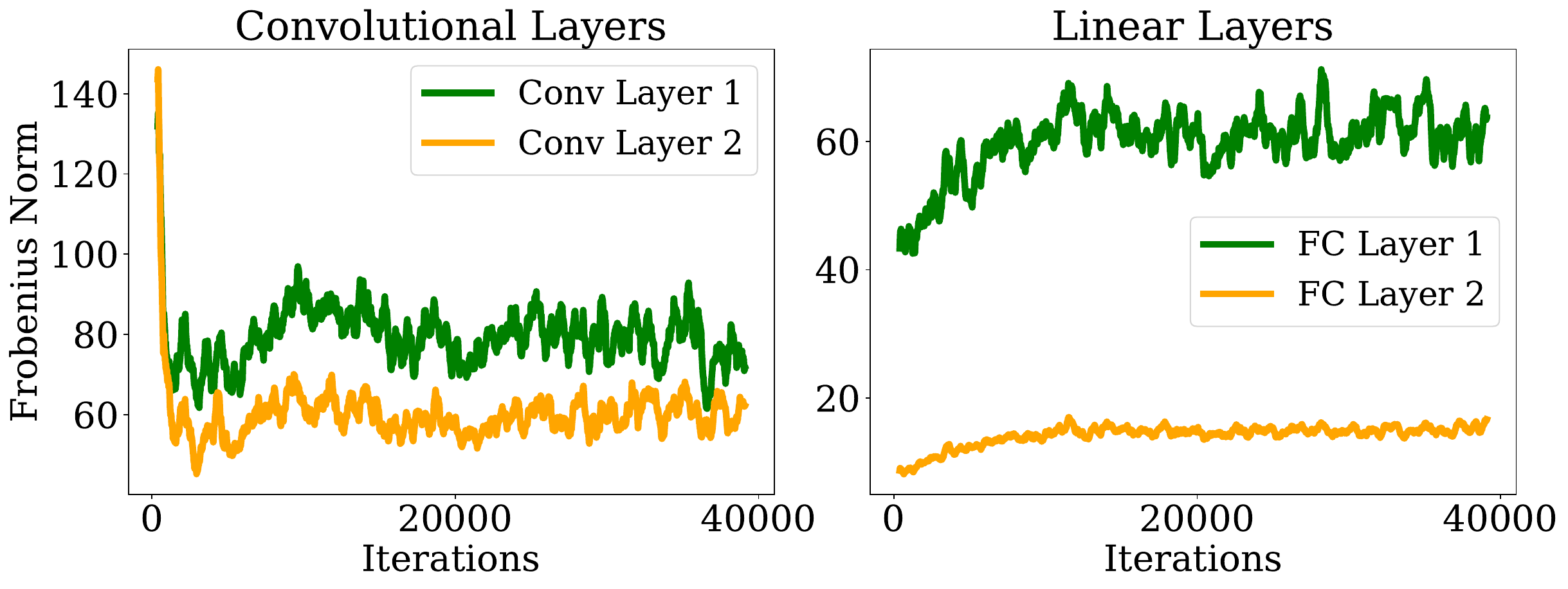}
    \caption{Trajectories of the Frobenius norm of the activation covariance for LeNet-5 layers on CIFAR-10 for 100 epochs.}
    \label{fig:fro_norm}
\end{figure}

In support of Assumption~\ref{as:bounded_actv}, empirical observations
indicate that the Frobenius norm of the activation covariance
consistently exhibits natural lower and upper bounds across
layers. Figure~\ref{fig:fro_norm} provides a toy example illustrating
the 
change of the Frobenius norm of the activation covariance at
each layer 
during the training.
We trained
LeNet-5~\cite{lecun98lenet5} on CIFAR-10 dataset using SGD for 100
epochs. The left subplot shows the Frobenius norms for the
convolutional layers, while the right subplot shows the Frobenius
norms for the fully connected layers. The norms are tracked across
training iterations. The 
result suggests that the
assumption of a bounded activation covariance norm is well-founded.


\begin{theorem}[Convergence of \nysact] \label{thm: convergence}
If the learning rate $\eta$ is set as $\eta = \mathcal{O}\left( \frac{(\lambda_{L} - \epsilon \nu_{L} + \rho)^2}{ \sigma^2 L(\lambda_{U}+ \epsilon \nu_{U} + \rho)}\right)$, then after $T$ iterations, the squared norm of gradient satisfies: 
\begin{align}
    \frac{1}{T} \sum_{k=0}^{T-1}&{\|\nabla\mathcal{L}(\mat{W}_{k})\|_F^2} \leq \frac{\E\mathcal{L}(\mat{W}_{0}) - \E\mathcal{L}(\mat{W}^{*})}{\eta T \left(  \frac{1}{\lambda_{U}+ \epsilon \nu_{U} + \rho} - \frac{\eta L }{2(\lambda_{L} - \epsilon \nu_{L} + \rho)^2} \right)} \nonumber \\
    &+  \frac{\eta \sigma^2 L }{2(\lambda_{L} - \epsilon \nu_{L} + \rho)^2 - \eta L (\lambda_{U}+ \epsilon \nu_{U} + \rho)}\,, \label{eq:convergence}
\end{align}
where $\mathcal{L}(\mat{W}^{*})$ denotes the global minimum of the loss function.
\end{theorem}
Theorem~\ref{thm: convergence} indicates that the convergence rate of \nysact is given by $\mathcal{O}(1/T)$. See Appendix~\ref{apdx:proof} for the proof.

\begin{remark}
  The convergence analysis 
  in \eqref{eq:convergence} demonstrates that the
  Nystr\"om approximation error $\epsilon$ directly impacts the
  convergence rate of \nysact. As the approximation error increases,
  convergence becomes slower, and conversely, reducing the approximation
  error accelerates convergence. Therefore, a crucial factor for
  achieving faster convergence lies in effectively controlling and
  minimizing the approximation error.  
\end{remark}


\section{Experiments}
\label{sec:experiment}
We assess the performance of \nysact on a range of image
classification tasks and compare it with other baseline methods. All
experiments were conducted using 2 Nvidia RTX6000 GPUs. 

\begin{table*}[tb]
\caption{Test accuracy (\%) of ResNet and DenseNet on CIFAR datasets}
\label{tab:cifar}
\centering
\begin{threeparttable}
\begin{tabular}{c|c|cc|cc|cc}
\toprule
\multirow{2}{*}{Dataset} & Model & \multicolumn{2}{c|}{ResNet-32} & \multicolumn{2}{c|}{ResNet-110} & \multicolumn{2}{c}{DenseNet-121} \\
& Epoch & 100 & 200 & 100 & 200 & 100 & 200 \\
\midrule
\multirow{9}{*}{CIFAR-10} & \textsc{SGD} & 92.80\scriptsize{$\pm$0.21} & 93.57\scriptsize{$\pm$0.29} & 93.30\scriptsize{$\pm$0.24} & 94.18\scriptsize{$\pm$0.43} & 95.33\scriptsize{$\pm$0.16} & 95.58\scriptsize{$\pm$0.13} \\
\cmidrule{2-8}
& \textsc{Adam} & 91.59\scriptsize{$\pm$0.09} & 92.28\scriptsize{$\pm$0.14} & 92.43\scriptsize{$\pm$0.08} & 92.90\scriptsize{$\pm$0.21} & 93.11\scriptsize{$\pm$0.19} & 93.35\scriptsize{$\pm$0.16} \\
& \textsc{AdamW} & 90.79\scriptsize{$\pm$0.16} & 91.83\scriptsize{$\pm$0.28} & 92.33\scriptsize{$\pm$0.25} & 93.19\scriptsize{$\pm$0.19} & 94.24\scriptsize{$\pm$0.10} & 94.56\scriptsize{$\pm$0.14} \\
\cmidrule{2-8}
& \textsc{KFAC} & 93.16\scriptsize{$\pm$0.17} & 93.78\scriptsize{$\pm$0.13} & 94.35\scriptsize{$\pm$0.13} & 94.64\scriptsize{$\pm$0.10} & 95.23\scriptsize{$\pm$0.16} & 95.57\scriptsize{$\pm$0.07} \\
& \textsc{Eva} & 93.07\scriptsize{$\pm$0.16} & 93.65\scriptsize{$\pm$0.14} & 94.18\scriptsize{$\pm$0.10} & 94.64\scriptsize{$\pm$0.09} & 95.30\scriptsize{$\pm$0.13} & 95.69\scriptsize{$\pm$0.11} \\
\cmidrule{2-8}
& \textsc{FOOF} & 93.61\scriptsize{$\pm$0.14}\footnotesize{$^{\dagger}$} & 94.05\scriptsize{$\pm$0.17}\footnotesize{$^{\dagger}$} & 94.70\scriptsize{$\pm$0.10}\footnotesize{$^{\dagger}$} & 95.09\scriptsize{$\pm$0.10}\footnotesize{$^{\dagger}$} & 95.79\scriptsize{$\pm$0.04}\footnotesize{$^{\dagger}$} & 95.95\scriptsize{$\pm$0.08}\footnotesize{$^{\dagger}$} \\
& \textsc{NysAct-g} & 93.12\scriptsize{$\pm$0.11} & 93.68\scriptsize{$\pm$0.21} & 94.48\scriptsize{$\pm$0.09} & 94.76\scriptsize{$\pm$0.12} & 95.53\scriptsize{$\pm$0.13} & 95.74\scriptsize{$\pm$0.10} \\
& \textsc{NysAct-s} & 93.28\scriptsize{$\pm$0.21}\footnotesize{$^{\ddagger}$} & 93.79\scriptsize{$\pm$0.22}\footnotesize{$^{\ddagger}$} & 94.53\scriptsize{$\pm$0.17}\footnotesize{$^{\ddagger}$} & 94.94\scriptsize{$\pm$0.16}\footnotesize{$^{\ddagger}$} & 95.60\scriptsize{$\pm$0.19}\footnotesize{$^{\ddagger}$} & 95.83\scriptsize{$\pm$0.08}\footnotesize{$^{\ddagger}$} \\
\midrule
\multirow{9}{*}{CIFAR-100} & \textsc{SGD} & 70.47\scriptsize{$\pm$0.38} & 70.67\scriptsize{$\pm$0.49} & 71.44\scriptsize{$\pm$1.90} & 72.48\scriptsize{$\pm$1.36} & 79.63\scriptsize{$\pm$0.15} & 80.32\scriptsize{$\pm$0.24} \\
\cmidrule{2-8}
& \textsc{Adam} & 67.16\scriptsize{$\pm$0.41} & 67.90\scriptsize{$\pm$0.55} & 70.10\scriptsize{$\pm$0.45} & 71.22\scriptsize{$\pm$0.44} & 73.48\scriptsize{$\pm$0.41} & 73.49\scriptsize{$\pm$0.21} \\
& \textsc{AdamW} & 65.23\scriptsize{$\pm$0.16} & 67.04\scriptsize{$\pm$1.08}  & 68.88\scriptsize{$\pm$0.31} & 70.59\scriptsize{$\pm$0.37} & 75.51\scriptsize{$\pm$0.23} & 76.30\scriptsize{$\pm$0.12} \\
\cmidrule{2-8}
& \textsc{KFAC} & 70.21\scriptsize{$\pm$0.34} & 70.91\scriptsize{$\pm$0.28}  & 73.10\scriptsize{$\pm$0.41} & 74.68\scriptsize{$\pm$0.33} & 79.79\scriptsize{$\pm$0.24} & 80.16\scriptsize{$\pm$0.10} \\
& \textsc{Eva} & 70.32\scriptsize{$\pm$0.31} & 71.11\scriptsize{$\pm$0.50}  & 73.55\scriptsize{$\pm$0.33} & 74.13\scriptsize{$\pm$0.34} & 79.32\scriptsize{$\pm$0.08} & 79.89\scriptsize{$\pm$0.27} \\
\cmidrule{2-8}
& \textsc{FOOF} & 71.21\scriptsize{$\pm$0.34}\footnotesize{$^{\dagger}$} & 71.82\scriptsize{$\pm$0.23}\footnotesize{$^{\dagger}$}  & 75.13\scriptsize{$\pm$0.26}\footnotesize{$^{\dagger}$} & 75.91\scriptsize{$\pm$0.31}\footnotesize{$^{\dagger}$} & 80.92\scriptsize{$\pm$0.28}\footnotesize{$^{\dagger}$} & 80.98\scriptsize{$\pm$0.25}\footnotesize{$^{\dagger}$} \\
& \textsc{NysAct-g} & 70.70\scriptsize{$\pm$0.18} & 71.14\scriptsize{$\pm$0.17}\footnotesize{$^{\ddagger}$} & 73.94\scriptsize{$\pm$0.38}\footnotesize{$^{\ddagger}$} & 74.70\scriptsize{$\pm$0.19} & 80.40\scriptsize{$\pm$0.24}\footnotesize{$^{\ddagger}$} & 80.70\scriptsize{$\pm$0.34}\footnotesize{$^{\ddagger}$} \\
& \textsc{NysAct-s} & 70.86\scriptsize{$\pm$0.44}\footnotesize{$^{\ddagger}$} & 71.12\scriptsize{$\pm$0.34} & 73.76\scriptsize{$\pm$0.45} & 75.01\scriptsize{$\pm$0.16}\footnotesize{$^{\ddagger}$} & 80.33\scriptsize{$\pm$0.29} & 80.54\scriptsize{$\pm$0.17} \\
\bottomrule
\end{tabular}
\begin{tablenotes}
\footnotesize
\item $\dagger$ and $\ddagger$ indicate the best and second-best test accuracies, respectively.
\end{tablenotes}
\end{threeparttable}
\end{table*}

\begin{figure}[tb]
    \centering
    \includegraphics[width=.6\linewidth]{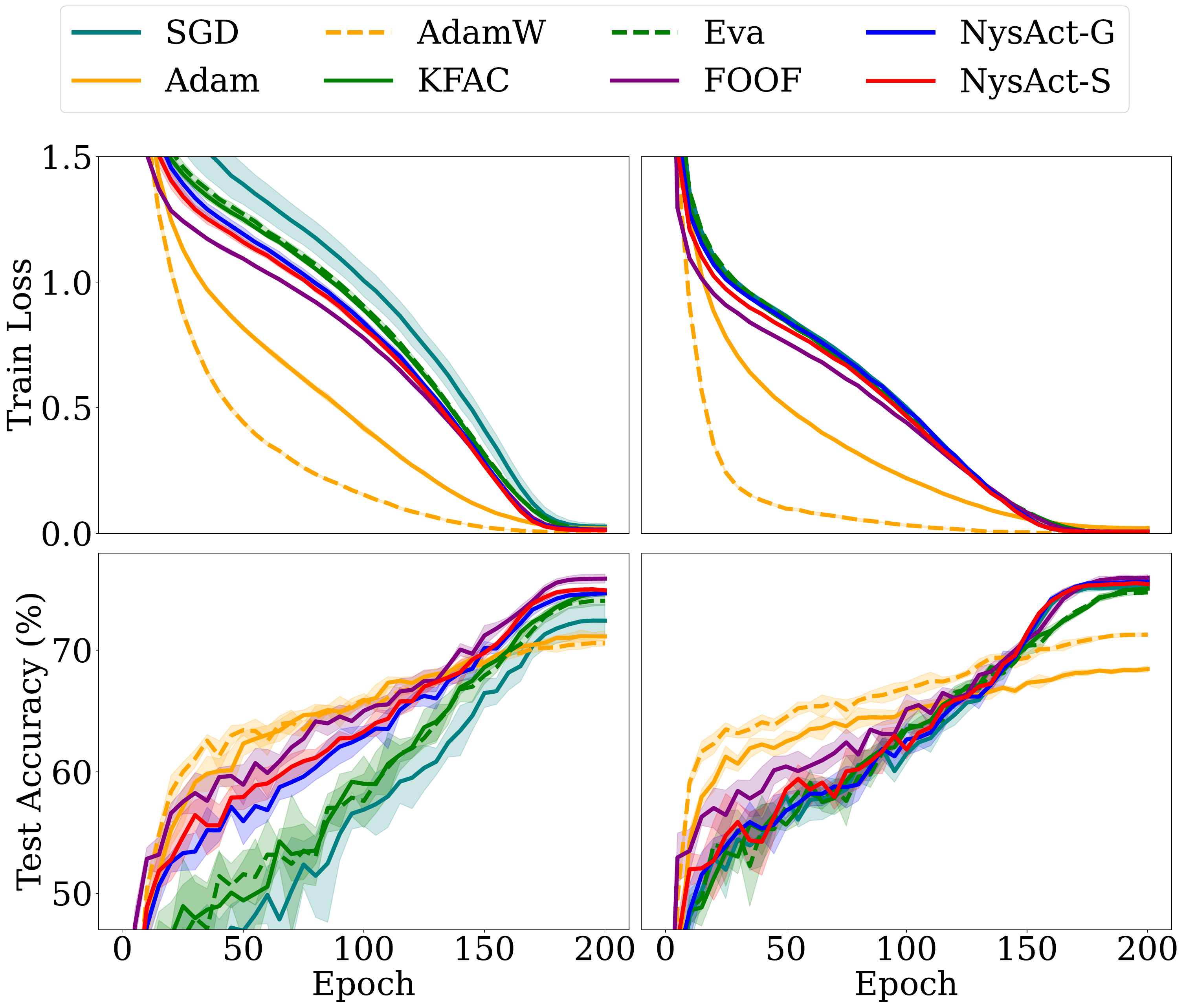}
    \caption{Comparison of training loss and test accuracy of optimizers on CIFAR-100 dataset using \textbf{(Left)} ResNet-110 and \textbf{(Right)} DenseNet-121.} 
    \label{fig:cifar100}
\end{figure}

\subsection{CIFAR Dataset}

\textbf{Settings. }
For the CIFAR datasets, we employed ResNet-32, ResNet-110, and
DenseNet-121~\cite{Huang2016DenselyCC}, training each model for 100
and 200 epochs. We used a mini-batch size of 128 and cosine annealing
learning rate scheduling~\cite{Loshchilov2016SGDRSG}. The reported
metrics in this section are averaged over 5 independent runs. We
compared \nysact against state-of-the-art first- and second-order
optimization methods. Specifically, we include SGD (with momentum) as an essential baseline and Adam and AdamW as adaptive first-order methods that precondition
gradients using their second moments. We evaluate KFAC and Eva as
second-order methods that precondition gradients using approximated
FIM. Finally, we include FOOF and \nysact as first-order methods that
employ activation covariance-based gradient preconditioning. 

\textbf{Training Results. }
The training results overall indicate that \nysact retains much of
FOOF's strong performance with minimal compromise. As shown in
Tables~\ref{tab:cifar}, \nysact outperforms most other baselines in
terms of test accuracy. While Adam and AdamW exhibit faster
convergence during the early stages of training, they ultimately
achieve lower test accuracy (i.e., relatively poor generalization
performance) compared to other methods. Second-order
methods generally outperform the first-order methods, such as SGD,
Adam, and AdamW. However, they show less effective performance in both
convergence rate and generalization compared to the activation
covariance-based preconditioning methods, FOOF and \nysact. When
comparing \nysact to FOOF, FOOF delivered the strongest results
overall, with \nysact following closely as a strong second. Given that
\nysact is designed as a scalable alternative to FOOF, this outcome
suggests that its approximation of the activation covariance matrix is
reasonably effective. In comparison to KFAC and Eva, \nysact either
matches or exceeds their performance in CIFAR-10 training, and it
distinctly outperforms all other methods in CIFAR-100 training across
all network architectures tested. 
Figure~\ref{fig:cifar100} shows the progression of training
loss and test accuracy over 200 epochs on CIFAR-100 dataset. The
results for ResNets clearly demonstrate that \nysact effectively
balances the fast convergence of first-order methods with the strong
generalization capabilities of second-order methods. For DenseNet-121,
\nysact performs comparably to the other second-order baselines. 
  
\begin{table}[tb]
\caption{Comparison of relative wall-clock time and memory usage over SGD on CIFAR datasets}
\label{tab:time_mem_cifar}
\centering
\begin{tabular}{c|cc|cc|cc}
\toprule
Model & \multicolumn{2}{c|}{ResNet-32} & \multicolumn{2}{c|}{ResNet-110} & \multicolumn{2}{c}{DenseNet-121} \\
(\# params) & \multicolumn{2}{c|}{(0.5M)} & \multicolumn{2}{c|}{(2M)} & \multicolumn{2}{c}{(8M)} \\
 & Time & Mem & Time & Mem & Time & Mem \\
\midrule
\textsc{KFAC} & 2.29$\times$ & 1.05$\times$ & 2.45$\times$ & 1.09$\times$ & 2.18$\times$ & 1.05$\times$ \\
\textsc{Eva} & 1.77$\times$ & 1.00$\times$ & 1.71$\times$ & 1.00$\times$ & 1.24$\times$ & 1.00$\times$ \\
\midrule
\textsc{FOOF} & 1.52$\times$ & 1.04$\times$ & 1.55$\times$ & 1.06$\times$ & 1.67$\times$ & 1.04$\times$ \\
\textsc{NysAct-g} & 1.33$\times$ & 1.00$\times$ & 1.36$\times$ & 1.00$\times$ & 1.22$\times$ & 1.00$\times$ \\
\textsc{NysAct-s} & 1.40$\times$ & 1.00$\times$ & 1.31$\times$ & 1.00$\times$ & 1.19$\times$ & 1.00$\times$ \\
\bottomrule
\end{tabular}
\end{table}

\textbf{Time and Memory Complexities. }
Table~\ref{tab:time_mem_cifar} highlights the computational efficiency
of \nysact. As demonstrated, \nysact 
is considerably faster
than both FOOF and KFAC, while at the same time  using less memory. Notably,
\nysact consistently achieves faster execution times compared to Eva,
which relies solely on vector multiplications during the computation
of preconditioners, across all tested architectures. 

\subsection{ImageNet Dataset}
\textbf{Settings. }
In our experiments on the ImageNet (ILSVRC 
2012)~\cite{deng2009imagenet} dataset, we trained ResNet-50 and DeiT
Small (DeiT-S)~\cite{Touvron2020TrainingDI} architectures for 100 and
200 epochs. Each training 
used a minibatch size of 1,024 and
employed the cosine learning rate decay. We evaluated \nysact against
the same 
baselines used in our CIFAR experiments. 

\begin{figure*}[tb]
  \centering
  \begin{subfigure}[t]{0.49\linewidth}
    \centering
    \includegraphics[width=1\textwidth]{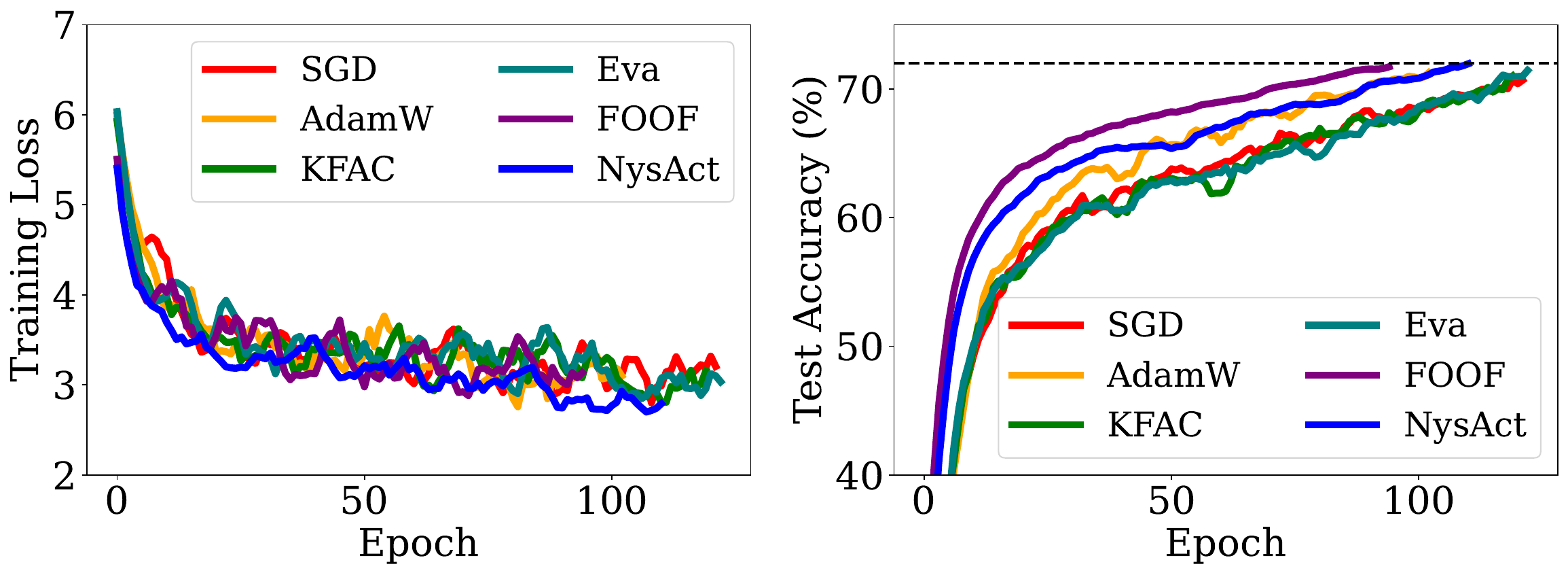}
    \caption{ResNet-50}
  \end{subfigure}
  \hfill
  \begin{subfigure}[t]{0.49\linewidth}
    \centering
    \includegraphics[width=1\textwidth]{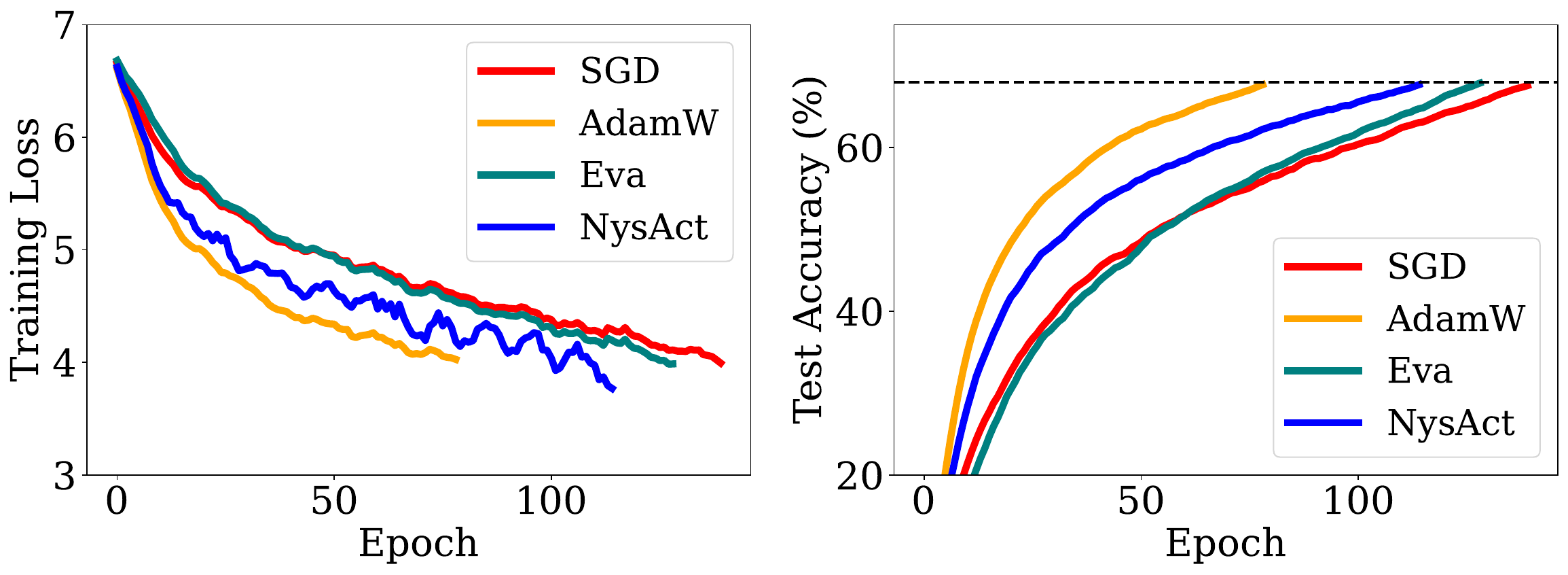}
    \caption{DeiT-S}
  \end{subfigure}
    \caption{Comparison of training loss and top-1 accuracy of optimizers on ImageNet dataset. The curves are truncated at thresholds corresponding to 90\% of the best top-1 accuracy achieved by SGD for each architecture.}
    \label{fig:imagenet_threshold}
\end{figure*}

\begin{table}[tb]
\caption{Top-1 accuracy (\%) of ResNet-50 and DeiT-S on ImageNet dataset}
\label{tab:imgnet_full} 
\centering
\begin{threeparttable}
\begin{tabular}{c|cc|cc}
\toprule
Model & \multicolumn{2}{c|}{ResNet-50} & \multicolumn{2}{c}{DeiT-S}  \\
Epoch & 100 & 200 & 100 & 200  \\
\midrule
\textsc{SGD} & 78.05 & 79.46 & 69.08 & 75.27  \\
\textsc{AdamW} & 76.73 & 79.14 & 73.78 & 77.96  \\
\midrule
\textsc{KFAC} & 78.16 & 79.34 & 69.84 & \xmark  \\
\textsc{Eva} & 77.71 & 79.48 & 69.67 & 76.57  \\
\midrule
\textsc{FOOF} & 78.37 & 79.69 & 65.37 & \xmark  \\
\textsc{NysAct-S} & 75.62 & 78.77 & 70.72 & 76.16 \\
\bottomrule
\end{tabular}
\begin{tablenotes}
\footnotesize
\item \xmark \ indicates a training failure.
\end{tablenotes}
\end{threeparttable}
\end{table}

\begin{table}[tb]
\caption{Number of epochs and wall-clock time required to achieve a threshold Top-1 accuracy for ResNet-50 (72\%) and DeiT-S (68\%) on ImageNet dataset}
\label{tab:imgnet_threshold} 
\centering
\begin{threeparttable}
\begin{tabular}{c|cc|cc}
\toprule
Model & \multicolumn{2}{c|}{ResNet-50} & \multicolumn{2}{c}{DeiT-S}  \\
 & Epochs & Time (hrs) & Epochs & Time (hrs) \\
\midrule
\textsc{SGD} & 126 & 8.48 & 144 & 9.61  \\
\textsc{AdamW} & 107 & 7.28 & 83 & 5.59  \\
\midrule
\textsc{KFAC} & 124 & 10.10 & \xmark & \xmark  \\
\textsc{Eva} & 127 & 8.72 & 133 & 8.96 \\
\midrule
\textsc{FOOF} & 99 & 8.33 & \xmark & \xmark  \\
\textsc{NysAct-S} & 115 & 8.21 & 119 & 8.18 \\
\bottomrule
\end{tabular}
\begin{tablenotes}
\footnotesize
\item \xmark \ indicates a training failure.
\end{tablenotes}
\end{threeparttable}
\end{table}

\textbf{Training Results. }
The experimental results on ImageNet dataset are summarized in
Table~\ref{tab:imgnet_full}, Table~\ref{tab:imgnet_threshold} and
Figure~\ref{fig:imagenet_threshold}.
For ResNet-50, \nysact
achieves slightly lower top-1 accuracy than other baselines like SGD,
AdamW, and FOOF.
However, in Figure~\ref{fig:imagenet_threshold}, we set the
threshold accuracy at 72\%, which corresponds to 90\% of the best
possible top-1 accuracy achievable by SGD in this setting. As shown,
\nysact requires fewer epochs
to achieve 72\% top-1 accuracy compared to SGD (115 epochs vs 126), as
well as second-order methods like KFAC (124 epochs) and Eva (127
epochs).
While AdamW and FOOF need slightly fewer
epochs to reach the threshold, \nysact still outperforms FOOF in terms
of wall-clock time. It achieves the second-best time of 8.21 hours,
following AdamW's 7.28 hours, making it highly efficient compared to
the 10.10 hours for KFAC and 8.72 hours for Eva. This indicates that
\nysact strikes a favorable balance between computational efficiency
and accuracy. 

For DeiT-S, where AdamW is known for its strong performance in terms
of convergence and accuracy, \nysact closely follows this
benchmark. \nysact achieves the second-best top-1 accuracy at 100
epochs and performs competitively at 200 epochs, trailing AdamW and
Eva. In terms of efficiency, \nysact requires 119 epochs to reach 68\%
top-1 accuracy (which represents 90\% of the best top-1 accuracy of
SGD in this architecture) compared to 144 epochs for SGD and 133
epochs for Eva. Additionally, \nysact’s wall-clock time is shorter
than that of both methods, making it the most efficient
preconditioning method in this setting.
On the other hand, KFAC and FOOF failed to train DeiT-S for 200 epochs due
to numerical issues in preconditioner inversion. \nysact’s strong performance
with DeiT-S underscores its scalability and adaptability to
attention-based architectures, even in a domain where AdamW is
typically dominant. 

\begin{table}[tb]
\caption{Comparison of relative wall-clock time and memory usage over SGD on ImageNet dataset}
\label{tab:time_mem_imagenet}
\centering
\begin{tabular}{c|cc|cc}
\toprule
Model (\# params)  & \multicolumn{2}{c|}{ResNet-50 (27M)} & \multicolumn{2}{c}{DeiT-S (22M)} \\
 & Time & Mem & Time & Mem  \\
\midrule
\textsc{KFAC} & 1.21$\times$ & 1.02$\times$ & 1.37$\times$ & 1.03$\times$  \\
\textsc{Eva} & 1.02$\times$ & 1.00$\times$ & 1.01$\times$ & 1.00$\times$  \\
\midrule
\textsc{FOOF} & 1.25$\times$ & 1.01$\times$ & 1.11$\times$ & 1.02$\times$  \\
\textsc{NysAct-S} & 1.06$\times$ & 1.00$\times$ & 1.03$\times$ & 1.00$\times$\\
\bottomrule
\end{tabular}
\end{table}

\textbf{Time and Memory Complexities. }
Table~\ref{tab:time_mem_imagenet} presents the efficiency of \nysact
on ImageNet dataset, demonstrating its superior scalability compared
to both FOOF and KFAC. Despite its faster execution,
\nysact maintains memory usage on par with SGD.
Additionally, \nysact achieves
comparable time and memory efficiency to Eva. This highlights
\nysact’s capability to balance speed and resource efficiency across
architectures, reinforcing its scalability and practicality for
large-scale training.


\section{Conclusion}
\label{sec:conclusion}
We introduced \nysact, a scalable 
stochastic preconditioned gradient method that effectively reduces the
computational complexity associated with activation covariance-based
preconditioning while maintaining a fast convergence rate and strong
generalization performance. Our extensive empirical evaluations on
image classification tasks demonstrate that \nysact significantly
improves end-to-end training time compared to other advanced
preconditioning methods
including KFAC, Eva, and FOOF.
Furthermore, \nysact delivers better test accuracy compared to first-order methods such as SGD and Adam(W). By addressing the limitations of both first- and second-order methods, \nysact offers an optimal blend between them, making it a scalable yet powerful optimization choice for deep learning tasks.


\bibliographystyle{plain}

\newpage
\appendix
\addcontentsline{toc}{section}{Appendix}

\section{Proof of Theorem~\ref{thm: convergence}} \label{apdx:proof}
The update rule of \nysact is given by
\begin{equation*}
    \mat{W}_{k+1} = \mat{W}_{k} - \eta\mat{G}(\mat{W}_k) (\mat{A}^{nys}_{k} + \rho \mat{I})^{-1}\,,
\end{equation*}
where $\mat{W}$ represents the weights in matrix form, $\mat{G}$ is the mini-batch gradient of loss $\mathcal{L}$ w.r.t. the weights in matrix form, and $\mat{A}^{nys}$ denotes the Nystr\"om approximated activation covariance. The Assumption~\ref{as:nystrom_error} gives the following inequalities, for all $i$:
\begin{equation}
    \lambda_{i}(\mat{A}) - \epsilon\|\mat{A}\|_F \leq \lambda_{i}(\mat{A}^{nys}) \leq \lambda_{i}(\mat{A}) + \epsilon\|\mat{A}\|_F. \label{eq:eigen_bounds}
\end{equation}

\begin{proof}
Using the Assumption~\ref{as:smoothness}, the Taylor expansion of the loss around $\mat{W}_k$ gives
\begin{equation*}
    \mathcal{L}(\mat{W}_{k+1}) \leq \mathcal{L}(\mat{W}_{k}) + \langle \nabla\mathcal{L}(\mat{W}_{k}), \mat{W}_{k+1} - \mat{W}_{k} \rangle + \frac{L}{2}\|\mat{W}_{k+1} - \mat{W}_{k}\|_{F}^2.
\end{equation*}
Substituting $\mat{W}_{k+1} - \mat{W}_{k} = - \eta\mat{G}(\mat{W}_k) (\mat{A}^{nys}_{k} + \rho \mat{I})^{-1}$, we have
\begin{align*}
    \mathcal{L}(\mat{W}_{k+1}) &\leq \mathcal{L}(\mat{W}_{k}) - \eta\langle \nabla\mathcal{L}(\mat{W}_{k}), \mat{G}(\mat{W}_k) (\mat{A}^{nys}_{k} + \rho \mat{I})^{-1} \rangle + \frac{\eta^2 L}{2}\|\mat{G}(\mat{W}_k) (\mat{A}^{nys}_{k} + \rho \mat{I})^{-1}\|_{F}^2  \\
    &= \mathcal{L}(\mat{W}_{k}) - \eta \Trace(\nabla\mathcal{L}(\mat{W}_{k}) (\mat{A}^{nys}_{k}+ \rho \mat{I})^{-1}\mat{G}(\mat{W}_k)^\intercal) + \frac{\eta^2 L}{2}\|\mat{G}(\mat{W}_k) (\mat{A}^{nys}_{k} + \rho \mat{I})^{-1}\|_{F}^2  \\
    \intertext{$\Trace(\mat{A}\mat{B}\mat{C}) \leq \|\mat{A}\|_F \|\mat{B}\mat{C}\|_F \leq \|\mat{A}\|_F\|\mat{B}\|_2 \|\mat{C}\|_F$, where $\|\cdot\|_2$ denotes the spectral norm, gives}
    &\leq \mathcal{L}(\mat{W}_{k}) - \eta \|\nabla\mathcal{L}(\mat{W}_{k})\|_F \|(\mat{A}^{nys}_{k} + \rho \mat{I})^{-1}\|_2 \|\mat{G}(\mat{W}_k)\|_F  + \frac{\eta^2 L }{2} \|\mat{G}(\mat{W}_k)\|_{F}^2 \|(\mat{A}^{nys}_{k} + \rho \mat{I})^{-1}\|_{F}^2 \\
    &\leq \mathcal{L}(\mat{W}_{k}) - \frac{\eta\|\nabla\mathcal{L}(\mat{W}_{k})\|_F\|\mat{G}(\mat{W}_k)\|_F }{\lambda_{\max}(\mat{A}^{nys}_{k}) + \rho}  + \frac{\eta^2 L \|\mat{G}(\mat{W}_k)\|_{F}^2}{2(\lambda_{\min}(\mat{A}^{nys}_{k}) + \rho)^2} \\
    \intertext{Assumption~\ref{as:bounded_actv} and \eqref{eq:eigen_bounds} yield}
    &\leq \mathcal{L}(\mat{W}_{k}) -  \frac{\eta\|\nabla\mathcal{L}(\mat{W}_{k})\|_F \|\mat{G}(\mat{W}_k)\|_F }{\lambda_{\max}(\mat{A}_{k}) + \epsilon \nu_{U} + \rho}  +  \frac{\eta^2 L\|\mat{G}(\mat{W}_k)\|_{F}^2}{2(\lambda_{\min}(\mat{A}_{k}) - \epsilon \nu_{L} + \rho)^2} \\
     &\leq \mathcal{L}(\mat{W}_{k})-  \frac{\eta\|\nabla\mathcal{L}(\mat{W}_{k})\|_F \|\mat{G}(\mat{W}_k)\|_F }{\lambda_{U} + \epsilon \nu_{U} + \rho}  + \frac{\eta^2 L \|\mat{G}(\mat{W}_k)\|_{F}^2}{2(\lambda_{L} - \epsilon \nu_{L} + \rho)^2}.
\end{align*}
Taking expectations over the training samples and using \ref{as:grad_prop}, we have
\begin{align}
    \E\mathcal{L}(\mat{W}_{k+1}) &\leq \E\mathcal{L}(\mat{W}_{k})-  \frac{\eta\|\nabla\mathcal{L}(\mat{W}_{k})\|_F^2 }{\lambda_{U} + \epsilon \nu_{U} + \rho}  + \frac{\eta^2 L \E\|\mat{G}(\mat{W}_k)\|_{F}^2}{2(\lambda_{L} - \epsilon \nu_{L} + \rho)^2} \nonumber \\
     &\leq \E\mathcal{L}(\mat{W}_{k})-  \underbracket{\frac{\eta\|\nabla\mathcal{L}(\mat{W}_{k})\|_F^2 }{\lambda_{U} + \epsilon \nu_{U} + \rho}}_{\circled{1}}  + \underbracket{\frac{\eta^2 L (\sigma^2 + \|\nabla\mathcal{L}(\mat{W}_{k})\|_F^2)}{2(\lambda_{L} - \epsilon \nu_{L} + \rho)^2}}_{\circled{2}}. \label{eq:loss_bound}
\end{align}
For the loss to decrease, it is necessary for the term $\circled{1}$ to dominate the term $\circled{2}$. This requires that the learning rate $\eta$ be chosen as
\begin{align}
    \eta &\leq \frac{2(\lambda_{L} - \epsilon \nu_{L} + \rho)^2 \|\nabla\mathcal{L}(\mat{W}_{k})\|_F^2}{ L(\lambda_{U}+ \epsilon \nu_{U} + \rho)(\sigma^2 + \|\nabla\mathcal{L}(\mat{W}_{k})\|_F^2)} \nonumber \\
    &\leq \frac{2(\lambda_{L} - \epsilon \nu_{L} + \rho)^2 C}{ L(\lambda_{U}+ \epsilon \nu_{U} + \rho)(\sigma^2 + C)}
    \label{eq:eta_condition}
\end{align}
Rearranging terms in \eqref{eq:loss_bound} and summing over $T$ iterations, we have
\begin{align*}
    \sum_{k=0}^{T-1}{\left(  \frac{\eta}{\lambda_{U}+ \epsilon \nu_{U} + \rho} - \frac{\eta^2 L }{2(\lambda_{L} - \epsilon \nu_{L} + \rho)^2} \right) \|\nabla\mathcal{L}(\mat{W}_{k})\|_F^2} &\leq \sum_{k=0}^{T-1}{\left(\E\mathcal{L}(\mat{W}_{k}) - \E\mathcal{L}(\mat{W}_{k+1}) + \frac{\eta^2 \sigma^2 L}{2(\lambda_{L} - \epsilon \nu_{L} + \rho)^2}\right)} \\
    &\leq \E\mathcal{L}(\mat{W}_{0}) - \E\mathcal{L}(\mat{W}_{T}) + \frac{\eta^2 \sigma^2 L T}{2(\lambda_{L} - \epsilon \nu_{L} + \rho)^2}
    \intertext{Assuming $\mathcal{L}(\mat{W}_{T}) \geq \mathcal{L}(\mat{W}^{*})$, where $\mathcal{L}(\mat{W}^{*})$ denotes the global minima of loss function, we have}
    &\leq \E\mathcal{L}(\mat{W}_{0}) - \E\mathcal{L}(\mat{W}^{*}) + \frac{\eta^2 \sigma^2 L T}{2(\lambda_{L} - \epsilon \nu_{L} + \rho)^2}
\end{align*}
Dividing by $T$ and rearranging terms, we obtain the following:
\begin{align*}
    \frac{1}{T} \sum_{k=0}^{T-1}{\|\nabla\mathcal{L}(\mat{W}_{k})\|_F^2} &\leq \frac{\E\mathcal{L}(\mat{W}_{0}) - \E\mathcal{L}(\mat{W}^{*})}{\eta T \left(  \frac{1}{\lambda_{U}+ \epsilon \nu_{U} + \rho} - \frac{\eta L }{2(\lambda_{L} - \epsilon \nu_{L} + \rho)^2} \right)} +  \frac{\eta \sigma^2 L }{2(\lambda_{L} - \epsilon \nu_{L} + \rho)^2 - \eta L (\lambda_{U}+ \epsilon \nu_{U} + \rho)}.
\end{align*}
\end{proof}

\section{Experimental Details}
\label{apdx:exp_detail}

\textbf{Hyperparameter Settings. }
\label{apdx:cifar_detail}
Table~\ref{tab:hyperpar_detail} outlines the hyperparameters utilized for training on CIFAR and ImageNet datasets. Here, $\eta$ refers to the learning rate, $\beta_{1}$ and $\beta_{2}$ represent the EMA coefficients for the first and second moments, and $\epsilon$ is a small constant added for numerical stability in Adam-based optimizers. In gradient preconditioning methods, $\beta_{2}$ corresponds to the EMA coefficient for preconditioner updates. The damping factor, $\rho$, controls the regularization of the covariance matrix, while $T_{cov}$ and $T_{inv}$ define the update frequencies for the covariance and inverse matrices, respectively. Lastly, $r$ denotes the approximation rank size utilized in \nysact. For CIFAR datasets, For KFAC and Eva, hyperparameter values followed the recommendations in \cite{Zhang2023EvaPS}. In contrast, for FOOF and \nysact, we conducted a grid search over learning rates [$0.001, 0.005, 0.01, 0.05, 0.1, 0.5$] and damping factors [$0.01, 0.05, 0.1, 0.5, 1, 5, 10$], while keeping the remaining settings consistent with other FIM-based preconditioners. 

For ImageNet dataset, we scaled up the learning rate $\eta$ by a factor of 5 for both SGD and preconditioning methods, using a mini-batch size of $1,024$, compared to the CIFAR training, while we decrease weight decay to $0.00002$. For both ResNets and DeiT, we tripled up the damping for KFAC and Eva to mitigate instability during preconditioner inversion. For all gradient preconditioning methods, we reduced the inversion frequency from 50 to 5, enabling the models to more frequently adjust to the changes in preconditioning matrices, particularly when training DeiT.

\begin{table}[h]
    \caption{Hyperparameter settings for CIFAR and ImageNet datasets training}
    \label{tab:hyperpar_detail}
    \centering
    \scriptsize
    \begin{tabular}{c|c|cccccccccc}
        \toprule
         Dataset & Optimizer & $\eta$ & Momentum & $\beta_{1}$ & $\beta_{2}$ & Weight decay & $\epsilon$ & $\rho$ & $T_{cov}$ & $T_{inv}$ & $r$ \\ 
         \midrule
         \multirow{6}{*}{CIFAR} & SGD & 0.1 & 0.9 & . & . &  0.0005 & . & . & . & . & .\\
         & Adam & 0.001 & . & 0.9 & 0.999 &  0.0005 & $1\times 10^{-8}$ & . & . & . & .\\
         & AdamW & 0.001 & . & 0.9 & 0.999 &  0.05 & $1\times 10^{-8}$ & . & . & . & .\\
         & KFAC & 0.1 & 0.9 & . & 0.95 & 0.0005 & . & 0.03 & 5 & 50 & .\\
         & Eva & 0.1 & 0.9 & . & 0.95 & 0.0005 & . & 0.03 & 5 & 50 & .\\
         & FOOF & 0.1 & 0.9 & . & 0.95 & 0.0005 & . & 1.0 & 5 & 50 & .\\
         & \nysact & 0.1 & 0.9 & . & 0.95 &  0.0005 & . & 1.0 & 5 & 50 & 10\\
         \midrule
         Dataset & Optimizer & $\eta$ & Momentum & $\beta_{1}$ & $\beta_{2}$ & Weight decay & $\epsilon$ & $\rho$ & $T_{cov}$ & $T_{inv}$ & $r$ \\ 
         \midrule
         \multirow{6}{*}{ImageNet} & SGD & 0.5 & 0.9 & . & . &  0.00002 & . &  . & . & . & .\\
         & AdamW & 0.001 & . & 0.9 & 0.999 &  0.05 & $1\times 10^{-8}$ & . & . & . & .\\
         & KFAC & 0.5 & 0.9 & . & 0.95 & 0.00002 & . &  0.1 & 5 & 50 / 5 & .\\
         & Eva & 0.5 & 0.9 & . & 0.95 & 0.00002 & . &  0.1 & 5 & 50 / 5 & .\\
         & FOOF & 0.5 & 0.9 & . & 0.95 & 0.00002 & . &  1.0 & 5 & 50 / 5 & .\\
         & \nysact & 0.5 & 0.9 & . & 0.95 &  0.00002 / 0.0001 & . &  1.0 & 5 & 50 / 5 & 50 / 20\\
         \bottomrule
    \end{tabular}
\end{table}

\textbf{Model Settings for ResNets and DeiT. }
\label{apdx:imagenet_detail}
Table~\ref{tab:imagenet_setting} summarizes the configurations used for training on the ImageNet dataset. Two architectures were explored: ResNet and DeiT-Small. For ResNet, we followed the PyTorch implementation~\cite{paszke2019pytorch}, and for DeiT, we applied the settings from \cite{Touvron2020TrainingDI}. Both models incorporated advanced training techniques such as Random Erasing~\cite{Zhong2017RandomED}, Label Smoothing~\cite{Szegedy2015RethinkingTI}, Mixup/CutMix~\cite{zhang2018mixup, Yun2019CutMixRS}, and Repeated Augmentation~\cite{Hoffer2019AugmentYB}. ResNet used TrivialAugment~\cite{Mller2021TrivialAugmentTY}, while DeiT employed RandAugment~\cite{Cubuk2019RandaugmentPA} and Stochastic Depth~\cite{Huang2016DeepNW}. Training was conducted at a resolution of 176 for ResNet and 224 for DeiT, with both models evaluated at a test resolution of 224. A mini-batch size of 1,024 was used with cosine learning rate decay and a 5-epoch warmup.

\begin{table}[h]
    \caption{Settings for ImageNet training}
    \label{tab:imagenet_setting}
    \centering
    \small
    \begin{tabular}{c|cc}
        \toprule
        Architecture & ResNets & DeiT \\
        \midrule
        Train Res & 176 & 224\\
        Test Res & 224 & 224 \\
        \midrule
        Batch size & 1,024 & 1,024 \\
        LR decay & cosine & cosine\\
        Warmup epochs & 5 & 5\\
        \midrule
        Label Smoothing & 0.1 & 0.1 \\
        Stochastic Depth & - & 0.2 \\
        Repeated Augmentation & \checkmark & \checkmark \\
        \midrule
        Horizontal flip & \checkmark  & \checkmark  \\
        Random Resized Crop & \checkmark  & \checkmark  \\
        Auto Augmentation & TrivialAugment & RandAugment(9/0.5) \\
        Mixup & 0.2 & 0.8\\
        Cutmix & 1.0 & 1.0 \\
        Random Erasing & 0.1 & 0.25 \\
        \bottomrule
    \end{tabular}
\end{table}

\newpage
\section{Hyperparameter Study}
\label{sec:ablation}
\begin{figure*}[tb]
  \centering
  \includegraphics[width=0.24\linewidth]{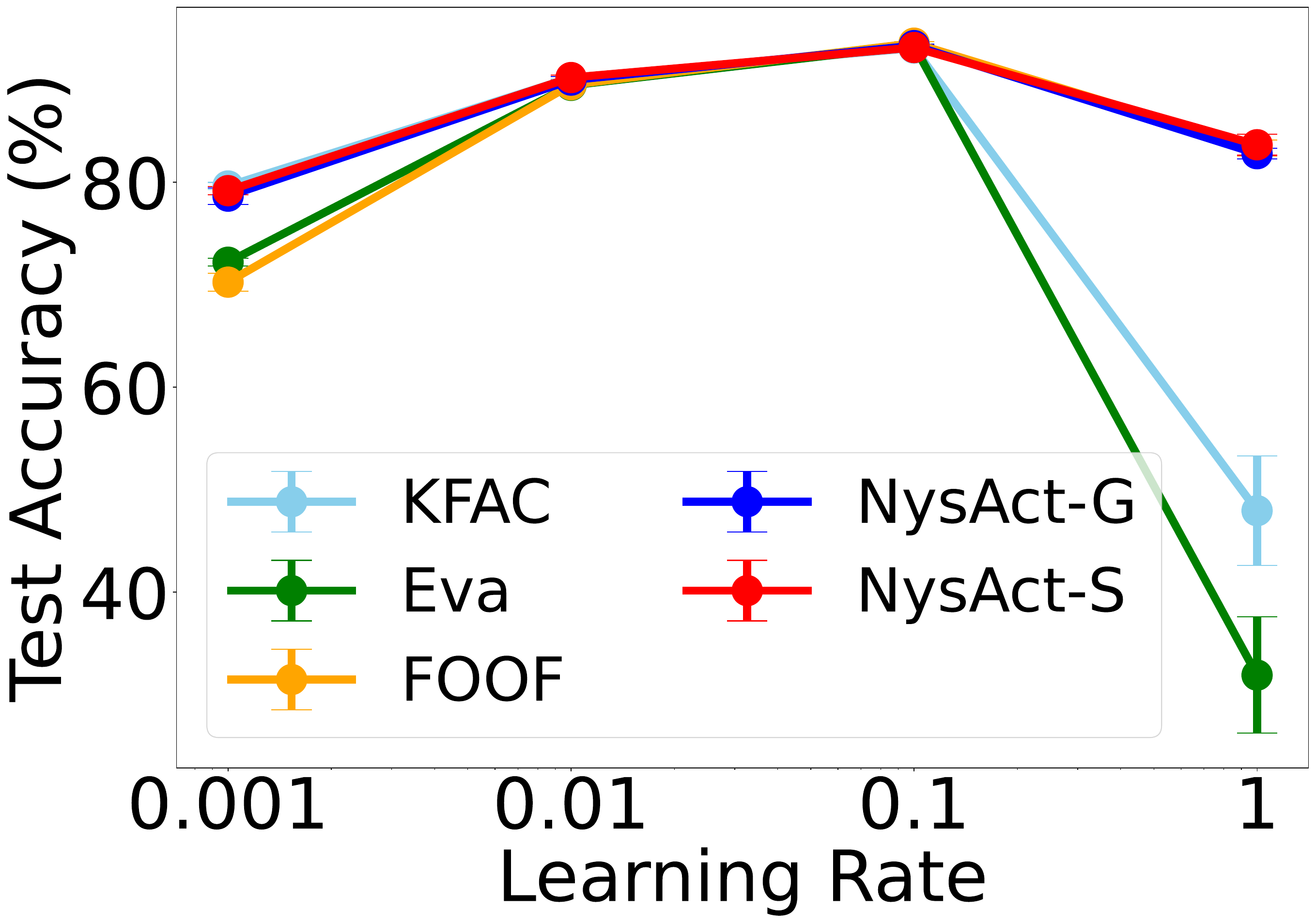}
  \includegraphics[width=0.24\linewidth]{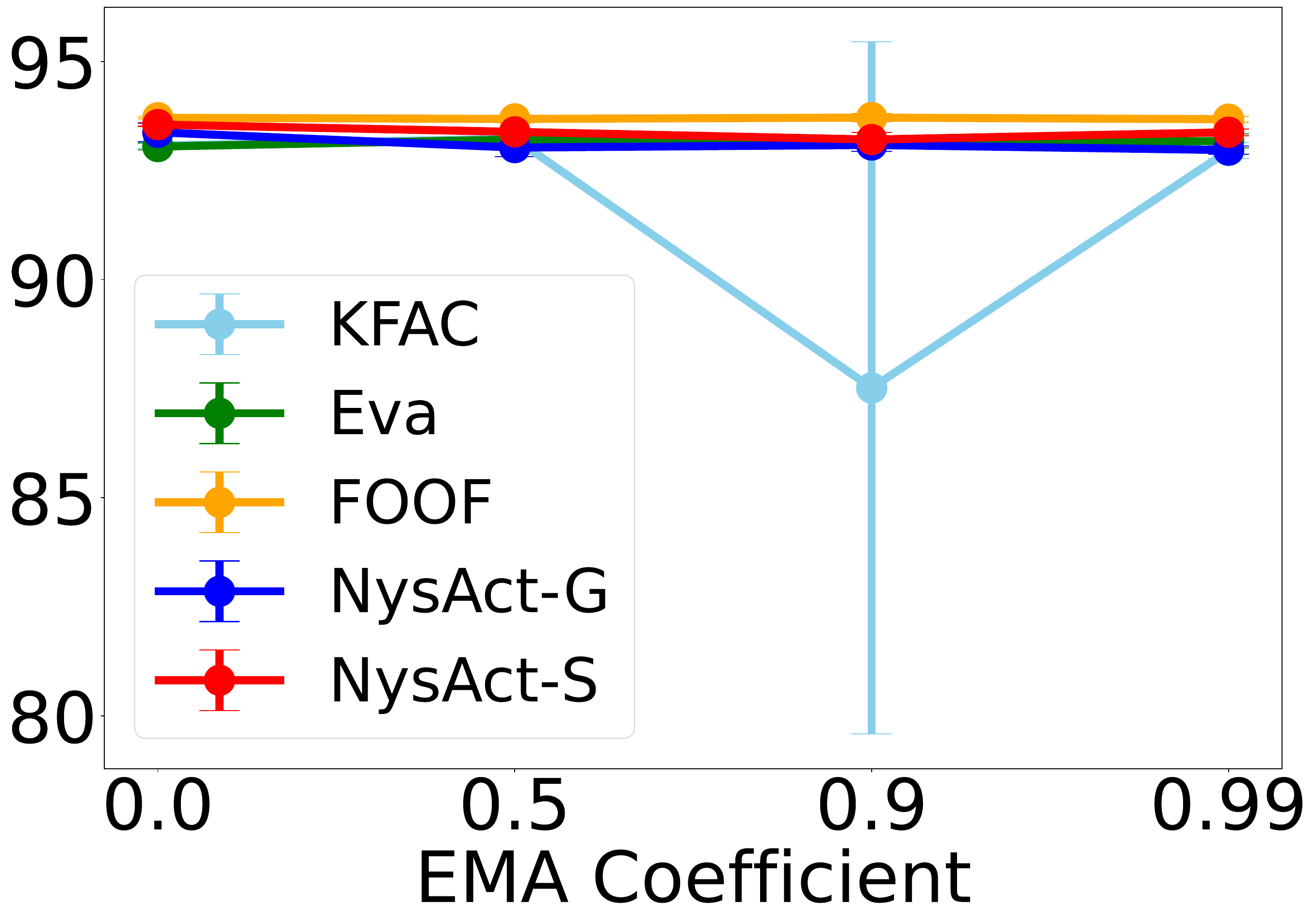}
  \includegraphics[width=0.24\linewidth]{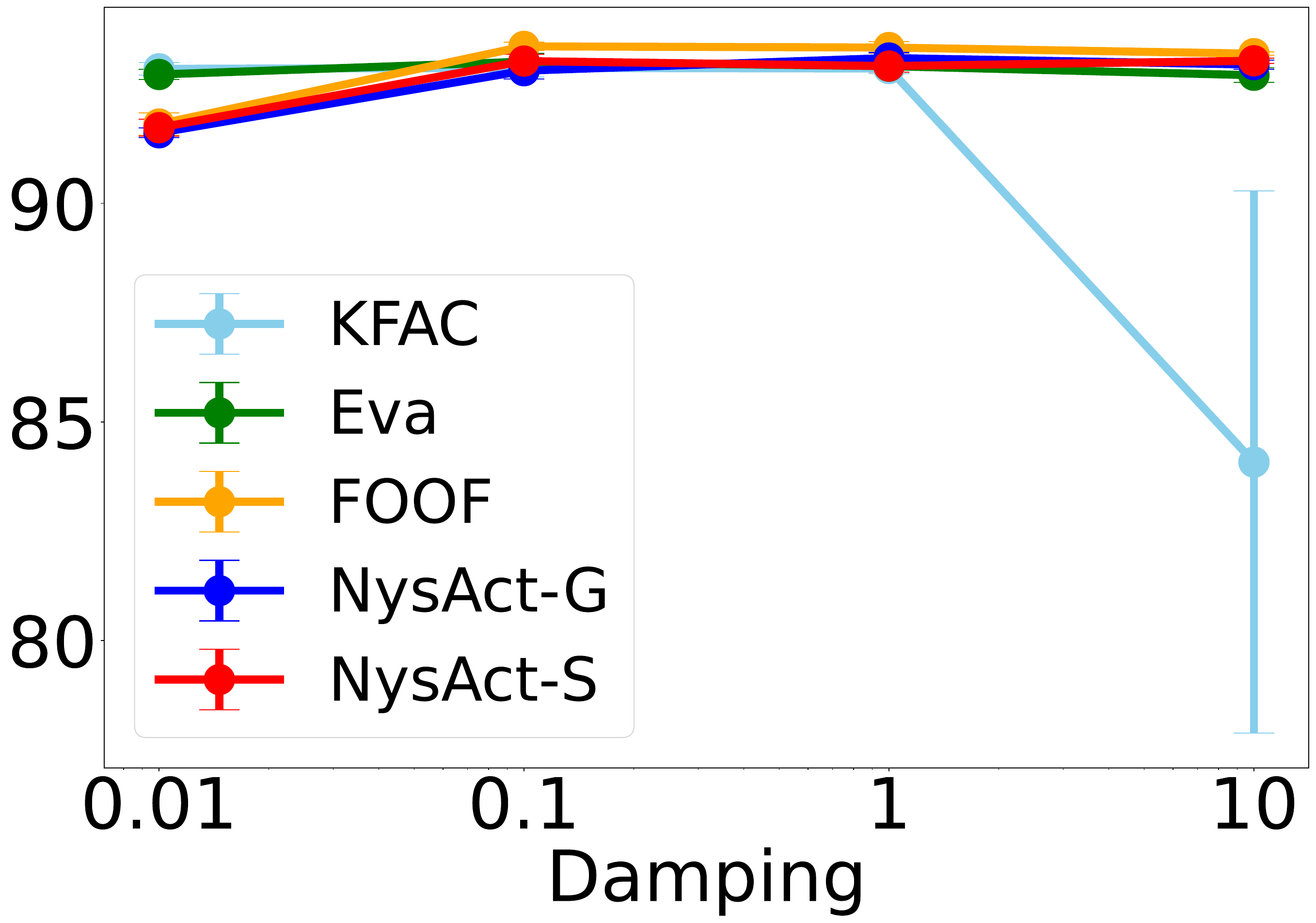}
  \includegraphics[width=0.24\linewidth]{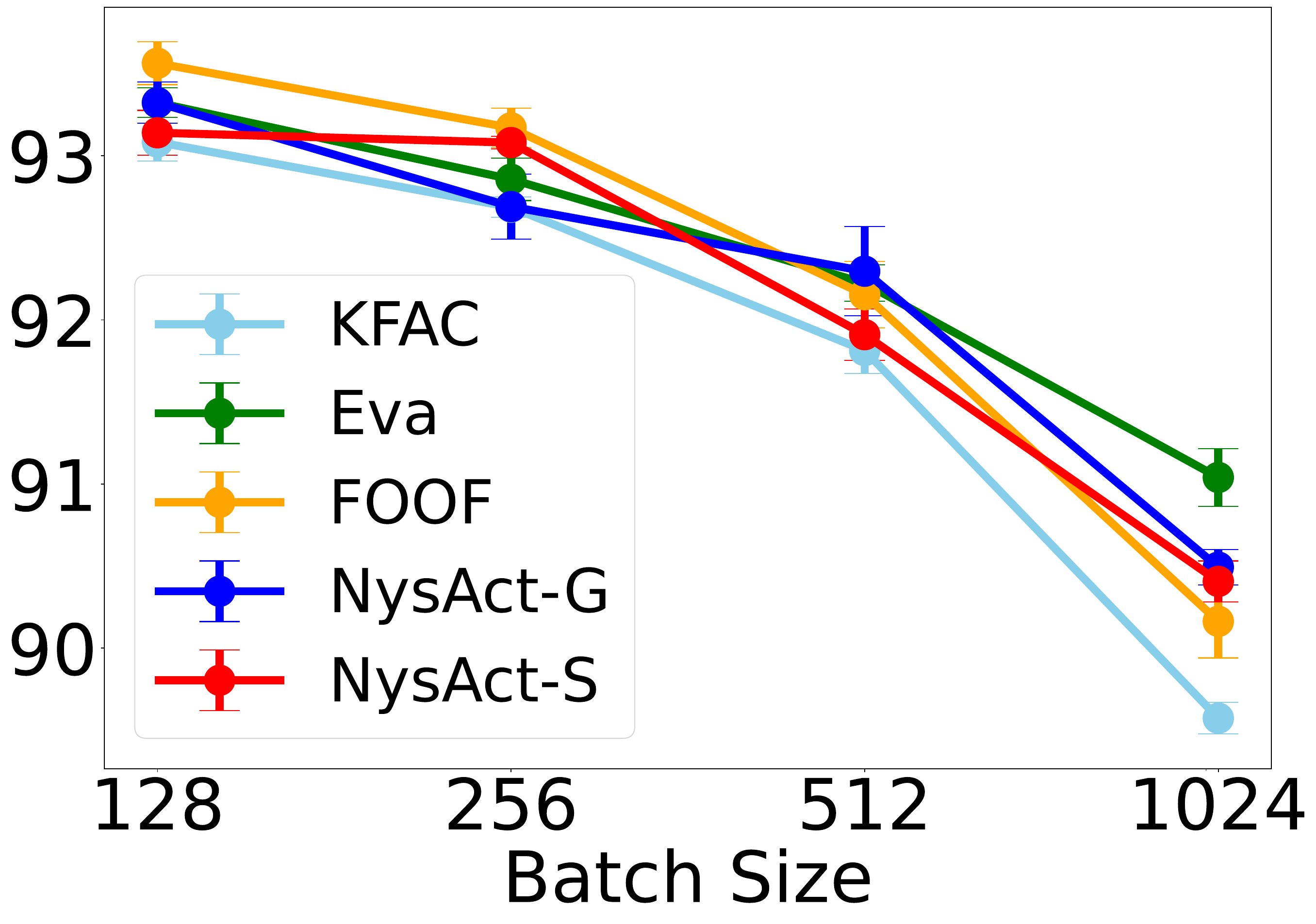}
  \caption{Comparison of the effects of learning rate, EMA coefficient, damping, and mini-batch size on gradient preconditioning methods, training ResNet-32 on CIFAR-10 for 100 epochs}
  \label{fig:ablation}
\end{figure*}

\begin{figure*}[tb]
  \centering
  \includegraphics[width=1\linewidth]{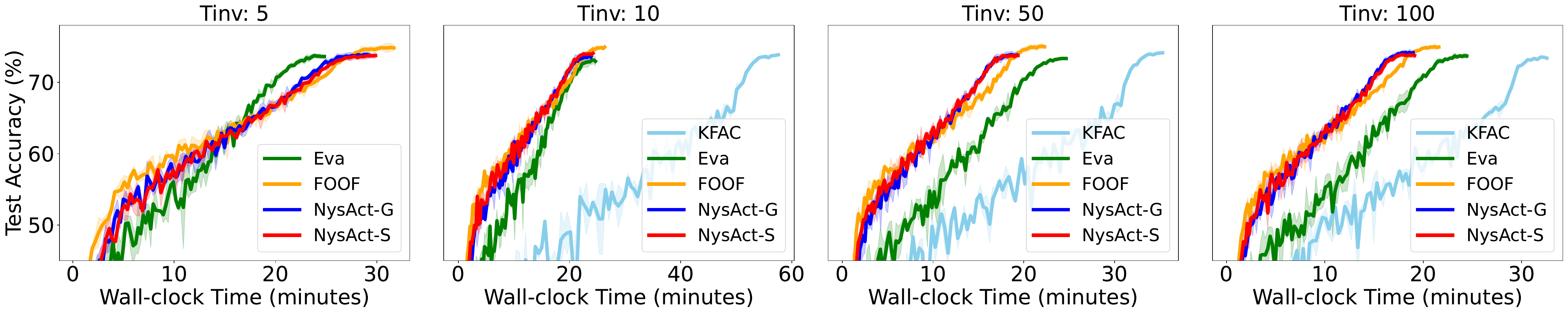}
  \caption{Comparison of wall-clock time for different inverse update frequencies during ResNet-110 training on the CIFAR-100 dataset}
  \label{fig:ablation_tinv}
\end{figure*}

\begin{figure*}[tb]
  \centering
  \begin{subfigure}[t]{0.49\linewidth}
    \centering
    \includegraphics[width=\linewidth]{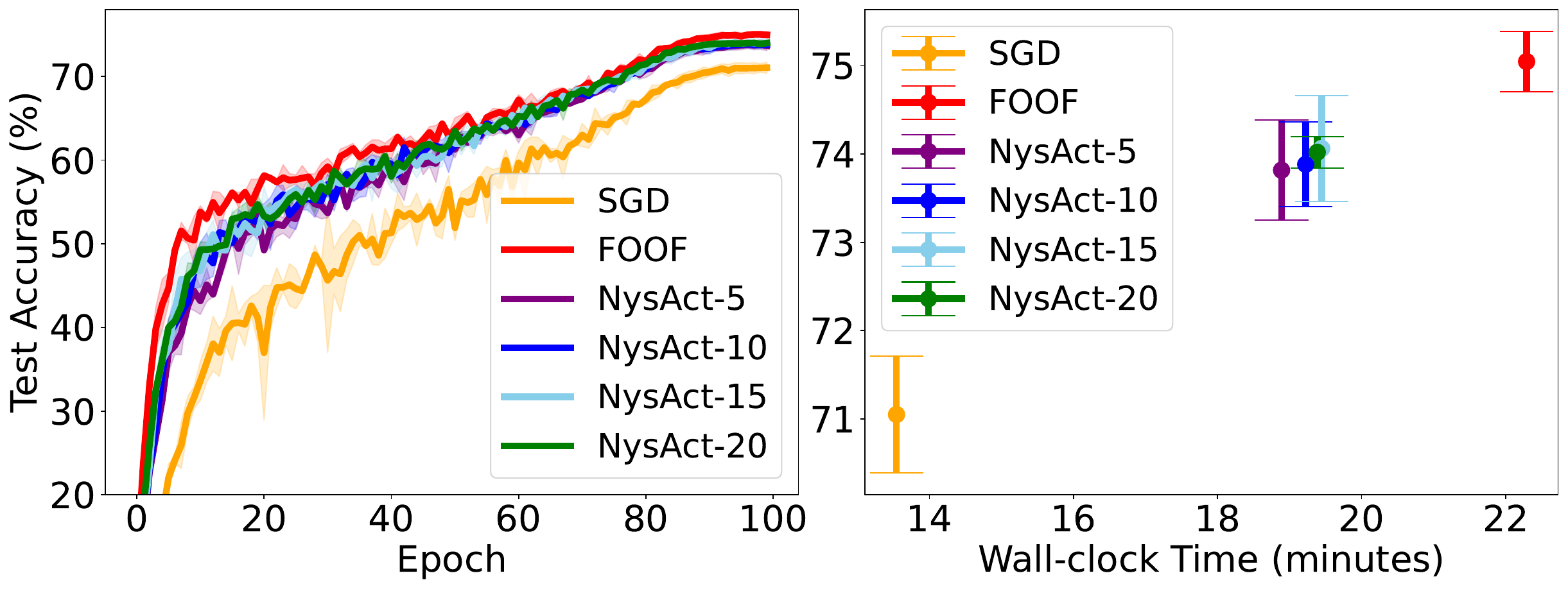}
    \caption{\textsc{NysAct}-Gaussian Sampling}
  \end{subfigure}
  \hfill
  \begin{subfigure}[t]{0.49\linewidth}
    \centering
    \includegraphics[width=\linewidth]{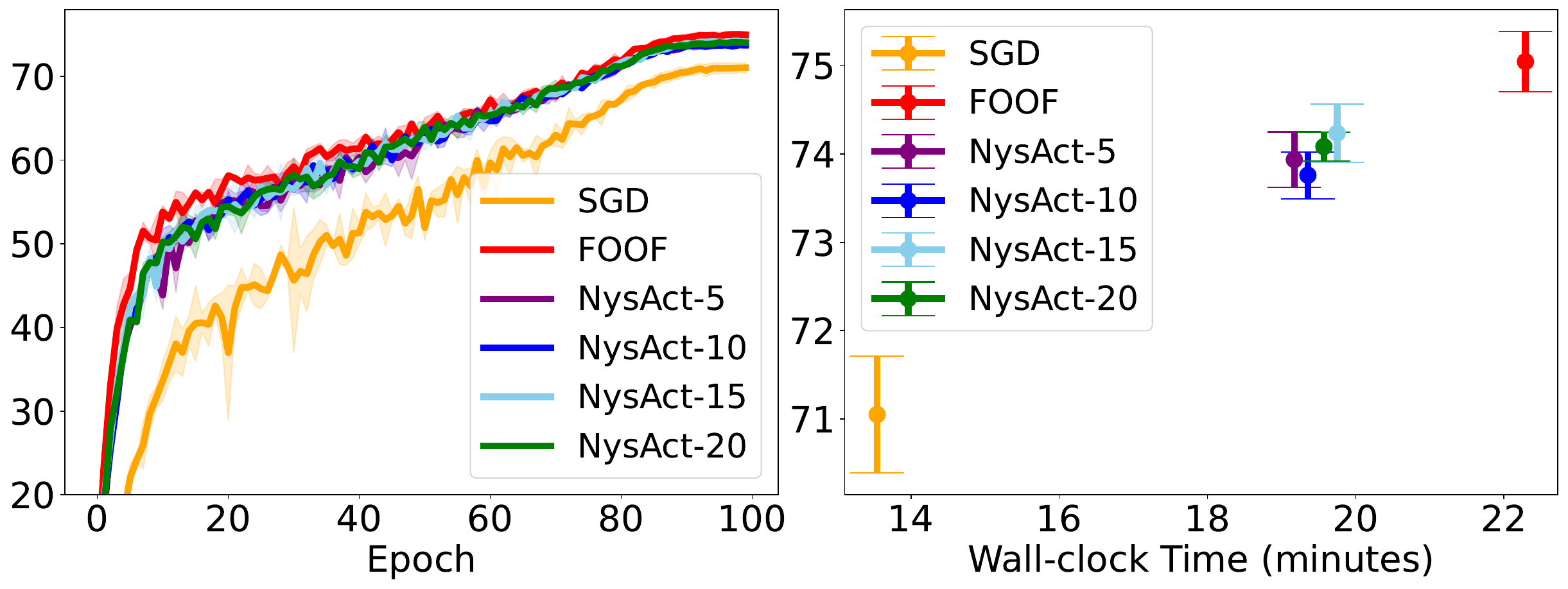}
    \caption{\textsc{NysAct}-Uniform Column Sampling}
  \end{subfigure}
  \caption{Comparison of test accuracy and wall-clock time for different ranks of \nysact during ResNet-110 training on CIFAR-100 dataset. The errorbar plots illustrate the comparison of test accuracy and wall-clock time for each method at their optimal epoch during the 100-epoch training period, along with the corresponding computational time in minutes.}
  \label{fig:ablation_rank}
\end{figure*}

We performed an ablation study on \nysact and compared its
results with other gradient preconditioning methods. We chose KFAC,
Eva, and FOOF as baselines because they all rely on approximations of
FIM and share similar hyperparameters, making them ideal for direct
comparison with \nysact. This analysis was carried out using ResNet-32
on CIFAR-10 and ResNet-110 on CIFAR-100, each trained for 100 epochs
across three different runs. 

\textbf{Essential Hyperparameters. }
Figure~\ref{fig:ablation} provides a comprehensive comparison of
\nysact with KFAC, Eva, and FOOF, focusing on essential hyperparameter
tuning. The first subplot highlights that all methods, including
\nysact, perform best at a learning rate of 0.1.
At the lower end of learning rate, i.e., when learning rate $\leq 0.01$,
Eva and FOOF struggle and show a drop in performance.
On the higher end, at the learning rate of 1.0, KFAC and
Eva experience significant performance degradation. In contrast,
\nysact maintains stable and consistent performance across the entire
range of learning rates tested. In the second subplot, \nysact, along
with the other methods, demonstrates consistent performance across
different EMA coefficients, with the exception of KFAC. This suggests
that \nysact's performance remains largely unaffected by changes of
the EMA coefficient, whereas KFAC exhibits noticeable fluctuations,
particularly at a coefficient value of 0.9. In the third subplot,
\nysact's test accuracy increases as 
the damping factor $\rho$ increases from 0.1 to 10.0.
At a damping value of 0.01, both FOOF and
\nysact experience a slight dip in performance relative to KFAC and
Eva. However, when the damping value reaches 10.0, KFAC displays
significant variability and a marked decline in performance. The
observation in the last subplot aligns with the broader trend in
optimization, where large-batch training often leads to degraded
network performance, as highlighted in previous
research~\cite{Goyal2017AccurateLM, Hoffer2017TrainLG}. Among the
methods compared, \nysact experiences a moderate decrease in test
accuracy as the batch size grows, showing a more stable performance
relative to other baselines.

\textbf{Inverse Update Frequencies. }
Figure~\ref{fig:ablation_tinv} shows
the effects of varying
inverse update frequencies for the preconditioning matrix, testing
intervals of 5, 10, 50, and 100 steps. The results suggest that
increasing the update frequency does not significantly compromise the
test accuracy of \nysact, while it contributes to reducing
computational overhead. For update frequencies of 10 steps or more,
\nysact achieved the fastest training time while maintaining the
second-best test accuracy, just behind FOOF.
At a frequency of 5
steps, Eva exhibited the fastest overall training time, with FOOF
being the slowest.
\nysact demonstrated a slightly faster training
time than FOOF. KFAC is absent from this subplot due to its frequent
failures in inverting the preconditioners. Notably, when $\tau_{inv}=50$ and $\tau_{inv}=100$
steps, Eva, despite being the lightest and most scalable variant of
KFAC, became slower than FOOF in this settings.

\textbf{Impact of Rank on \nysact. }
Figure~\ref{fig:ablation_rank} presents the impact of the rank
hyperparameter in \nysact. The subplots on the left display the
results for \textsc{NysAct-G}, while those on the right show the
results for \textsc{NysAct-S}. In both cases, \nysact outperforms SGD
in test accuracy and closely follows FOOF. When comparing the
sketching methods, Gaussian sampling exhibits larger variability in
test accuracy compared to subcolumn sampling, though both methods
achieve similar performance, around 74\% test accuracy. As the rank
increases, there is a subtle trend of improved test accuracy for
both sampling methods, aligning with the expectation that higher-rank
approximations better capture the original matrix's properties. The
findings suggest that \nysact effectively approximates the exact
activation covariance matrix with low ranks, as evidenced by the
minimal difference in test accuracy between rank-5 and rank-20
approximations, with overlapping error bars indicating negligible
variance. 

\end{document}